\documentclass{article}

\usepackage[nonatbib]{neurips_2020}
\usepackage[utf8]{inputenc}

\usepackage[square,numbers]{natbib}
\usepackage{cite}
\usepackage{amsmath,amssymb,amsfonts}
\usepackage{graphicx}
\usepackage{textcomp}
\usepackage{xcolor}
\usepackage{calc}
\usepackage{url}
\usepackage{textcomp}
\usepackage{multirow}
\usepackage{paralist}
\usepackage[utf8]{inputenc}
\usepackage{wrapfig}
\usepackage{subfigure}
\usepackage[abs]{overpic}
\usepackage{calc}
\usepackage{algpseudocode}
\usepackage{algorithm} 
\usepackage{pifont}
\DeclareUnicodeCharacter{2217}{*}
\DeclareMathOperator*{\argmax}{arg\,max}
\DeclareMathOperator*{\argmin}{arg\,min}

\usepackage[utf8]{inputenc} 
\usepackage[T1]{fontenc}    
\usepackage{hyperref}       
\usepackage{url}            
\usepackage{booktabs}       
\usepackage{nicefrac}       
\usepackage{microtype}

\title{Batch-Constrained Distributional Reinforcement Learning for Session-based Recommendation}

\author{
  Diksha Garg \\
  TCS Research\\
  New Delhi, India \\
  \texttt{diksha.7@tcs.com} \\   
   \And
   Priyanka Gupta \\
   TCS Research \\
   New Delhi, India \\
   \texttt{priyanka.g35@tcs.com} \\
   \AND
   Pankaj Malhotra \\
   TCS Research\\
   New Delhi, India \\
   \texttt{malhotra.pankaj@tcs.com} \\
   \And
   Lovekesh Vig \\
   TCS Research \\
   New Delhi, India \\
   \texttt{lovekesh.vig@tcs.com} \\
   \And
   Gautam Shroff \\
   TCS Research \\
   New Delhi, India \\
   \texttt{gautam.shroff@tcs.com} \\
}

\begin{document}
\nolinenumbers

\maketitle

\begin{abstract}
Most of the existing deep reinforcement learning (RL) approaches for session-based recommendations either rely on costly online interactions with real users, or rely on potentially biased rule-based or data-driven user-behavior models for learning.
In this work, we instead focus on learning recommendation policies in the pure batch or offline setting, i.e. learning policies solely from offline historical interaction logs or batch data generated from an unknown and sub-optimal behavior policy, without further access to data from the real-world or user-behavior models.
We propose \textit{BCD4Rec}: Batch-Constrained Distributional RL for Session-based Recommendations.
BCD4Rec builds upon the recent advances in batch (offline) RL and distributional RL to
learn from offline logs while dealing with the intrinsically stochastic nature of rewards from the users due to varied latent interest preferences (environments).
We demonstrate that BCD4Rec significantly improves upon the behavior policy as well as strong RL and non-RL baselines in the batch setting in terms of standard performance metrics like Click Through Rates or Buy Rates. 
Other useful properties of BCD4Rec include: 
i. recommending items from the correct latent categories indicating better value estimates despite large action space (of the order of number of items), and
ii. overcoming popularity bias in clicked or bought items typically present in the offline logs.

\end{abstract}

\section{Introduction}

Approaches for Session-based Recommendation (SR) aim to dynamically recommend items to a user based on the sequence of ongoing interactions (e.g. item clicks) in a session.
Several existing deep learning (DL) approaches for SR are designed to maximize the immediate (short-term) reward for recommendations, e.g. \citep{hidasi2018recurrent,liu2018stamp,wu2018session}.
More recently, deep reinforcement learning (DRL) approaches have been proposed that maximize the expected long-term cumulative reward by looking beyond the immediate user recommendation, e.g. \citep{chen2018generative,bai2019model,zhao2018deep}. Such approaches can, for instance, optimize recommendations for long-term user engagement instead of maintaining a myopic objective of optimizing the immediate user recommendation.

Model-free RL approaches for SR, e.g. \citep{zheng2018drn,zhao2018recommendations,zhao2018deep} rely on large-scale data collection by interacting with a population of users \citep{chen2018generative}. 
These interactions can be potentially costly as the initial recommendations from an untrained RL agent may leave a poor impression on users, and can lead to churn.
On the other hand, model-based RL approaches, e.g. \citep{chen2018generative,bai2019model,zhao2017deep}, rely on learning user-feedback simulation models that probe for rewards in previously unexplored regions of the state and action space. These approaches are sample-efficient in comparison to model-free approaches but have to rely on user-behavior model approximation from inherently biased logs \citep{bai2019model,yang2018unbiased}.

An alternative to the aforementioned approaches is learning an RL-based recommendation agent solely from historical logs obtained from a different sub-optimal recommendation policy. 
Once a batch of data from potentially sub-optimal policies is gathered, an RL agent can be learned from this fixed dataset overcoming the need for further feedback from the costly real-world interactions, or the need for often biased data-driven user-behavior models. 
Data can be collected by deploying less costly, easy-to-train, and fast-to-deploy heuristics- or rule-driven sub-optimal behavior policies, e.g. those based on k-nearest-neighbors \citep{jannach2017recurrent,garg2019sequence}, and then further used to improve upon the behavior policies.
Recently, several DRL approaches have been proposed for such a \textit{Batch RL} (a.k.a. Offline RL) setting where the agent is trained from a batch of data without access to further interactions, e.g. \citep{fujimoto2018off,fujimoto2019benchmarking,kumar2019stabilizing,levine2020offline}.

In this work, we study some of the SR-specific challenges in model-free batch RL approaches, and propose ways to mitigate them.
For instance, logs from sub-optimal policies have rewards for only a sparse set of state-action pairs, while is known to cause overestimation bias or errors in Q-learning \citep{fujimoto2018off,fujimoto2019benchmarking}.
This bias can be particularly severe in the SR setting where the action space can be very large (of the order of the number of items in the catalog, e.g. 1000s).
Furthermore, each user acts as a different version of the environment for the RL agent, lending intrinsic stochasticity to the environment \citep{bellemare2017distributional}. 
This stochasticity is even more apparent in the SR setting, where no past information or demographic details of the user (environment) are available. 
The effects of this stochasticity are amplified in the batch RL setting, where logs from sub-optimal policies are biased \citep{yang2018unbiased,bai2019model}, and do not depict the true user behavior characteristics.
For these reasons, robust estimation of the reward distribution from the environment (user) can be challenging in the batch learning scenario. 
Addressing these challenges in the batch learning setup for SR, we propose Batch-Constrained Distributional RL for Session-based Recommendations, or \textit{BCD4Rec}. 

The key contributions of this work can be summarized as follows:

    1. Inline with recent results from other domains \citep{fujimoto2019benchmarking,kumar2019stabilizing}, we first observe that the standard off-policy Q-learning approaches such as Deep Q-Networks (DQN) suffer in the batch RL setting for SR as well. In most cases, deep Q-learning using Deep Q Networks (DQN, \citep{mnih2013playing,van2016deep}) and other state-of-the-art non-RL approaches fail to improve upon the behavior policy from which the batch data was generated in the first place. We observe that these approaches can, at best, mimic the behavior policy.
    \newline
    2. We propose BCD4Rec: an approach for SR that is suited for the batch-constrained deep RL setup. 
    We adapt and build-upon the recent advancements in Batch RL \citep{fujimoto2018off,fujimoto2019benchmarking} and Distributional RL \citep{bellemare2017distributional,dabney2017distributional} (detailed in Section \ref{sec:approach_iqn}), and extend them for large state and action spaces frequently encountered in SR by using state and action embeddings. 
    \newline
    3. Through empirical evaluation on a real-world and a simulated dataset, we demonstrate that BCD4Rec can significantly improve upon the performance of the behavior policy as well as other strong RL and non-RL baselines. BCD4Rec further depicts several desirable properties: i. reducing overestimation error as measured by its ability to suggest meaningful items from the correct latent categories, 
    and ii. overcoming the bias in the offline logs and recommending relevant items with reduced popularity bias. 

\section{Related Work}
In recent years, heuristics-driven nearest neighbor-based approaches (e.g. \citep{jannach2017recurrent,ludewig2018evaluation,garg2019sequence}) as well as several deep learning approaches based on recurrent neural networks (RNNs) (e.g. \citep{hidasi2018recurrent}), graph neural networks (e.g. \citep{wu2018session,gupta2019niser}), attention networks (e.g. \citep{liu2018stamp}), etc. have been proposed for SR. 
The DL approaches provide state-of-the-art performance in next-step interaction prediction task but are myopic in their recommendations and do not take longer-term goals into account.  Several RL-based approaches based on MDPs \citep{shani2005mdp}, factored MDPs \citep{tavakol2014factored}, and approximate deep RL \citep{zhao2018deep,zhao2017deep,zhao2018recommendations} have been proposed for SR. These methods aim to optimize the long-term cumulative reward from users instead of next-step prediction tasks. 
However, these methods have been primarily proposed for online RL settings, and require costly experience collection by interacting with a population of users.
As we show empirically, vanilla deep Q-learning methods, as used in the above approaches, struggle in the batch RL settings. 

Several approaches for batch RL have been proposed to explore data-efficient RL approaches in absence of access to the real environment, e.g. \citep{sutton2016emphatic,fujimoto2018off,fujimoto2019benchmarking,jaques2019way,kumar2019stabilizing,levine2020offline,agarwal2020optimistic}. A promising approach to avoid overestimation error is to estimate (mimic) the unknown behavior policy, and use it to guide and constrain the action space of the RL agent \citep{kumar2019stabilizing,fujimoto2018off,fujimoto2019benchmarking}.
Very recently, these ideas have been used in deep Q-learning based batch RL for recommender systems in Generator Constrained Q-Learning (GCQ) \citep{wang2020off}. However, they explicitly leverage the user information which is not available in the SR setup considered in this work. Our work can be seen as an extension of GCQ to the SR setup. Moreover, GCQ considers constraining the action space of the RL agent by using item-frequency based approach. As we show in this work, frequency-based approaches are prone to popularity bias \citep{yang2018unbiased}, and can struggle to recommend relevant long-tail items. Furthermore, having access to user history, GCQ does not address the stochastic nature of user-behavior which can be critical in the SR setting.
To the best of our knowledge, this is the first study to demonstrate the advantage of using distributional RL \citep{dabney2017distributional,bellemare2017distributional} in the SR setting.

Many algorithms for off-policy evaluation from logged bandit feedback that utilize ideas from importance sampling, inverse propensity scoring, and counterfactual risk minimization have been proposed  \citep{li2011unbiased,swaminathan2017off}. 
However, these approaches have not been considered for session-based recommendations which involve large action spaces and sequential decision-making.
In this work, we look at the batch learning problem for session-based recommendations within the RL setup.

\section{BCD4Rec\label{sec:approach_iqn}}   

Consider a Markov Decision Process (MDP) \citep{sutton2011reinforcement} defined by the tuple of five elements ($\mathcal{S, A}, P, R, \gamma$), where $\mathcal{S}$ is the state space, $\mathcal{A}$ is the action space, $P(s'|s,a)$ is the transition probability from state $s$ to $s'$, $R(s,a)$ is the random variable reward function, $\gamma \in (0,1)$ is the discount factor, $s,s' \in \mathcal{S}$ and $a \in \mathcal{A}$. Given a policy $\pi$, the value function for the agent following the policy is given by the expected return of the agent $Q^{\pi}(s,a) := \mathbb{E} \left[Z^\pi(s,a)\right]= \mathbb{E}_{\pi}\left [\sum_{t=0}^{\infty}\gamma^t R(s_t,a_t)\right ],
$ where $s_t\sim P(\cdot|s_{t-1},a_{t-1})$, $a_t \sim \pi(\cdot|s_t)$, $s_0=s$, $a_0=a$.

The recommender agent (RA) in the SR setting interacts with a user (environment) by sequentially choosing the impression list of items (or the slate) to be recommended over a sequence of time steps, so as to maximize its cumulative reward while receiving feedback from the user. The state $s= \{s^1,s^2,\ldots,s^L\} \in \mathcal{S}$ corresponds to the  browsing history of the user consisting of the most recent $L$ interactions in the current session.
An action $a=\{a^1, a^2, \ldots, a^l\} \in \mathcal{A}$ corresponds to a slate or impression list of $l$ items chosen by the agent as a recommendation to the user based on the current state $s$, from a set $\mathcal{I}$ of currently available items not clicked by the user previously.
The transition probability $P(s'|s,a)$ from the current state $s$ to the next state $s'$ depends on the response of the user to the action $a$ taken by the RA in state $s$.
The immediate reward $r$ given state $s$ and action $a$ is determined by the response of the user, e.g. a click on an item results in $r=1$ while a skip results in $r=0$.
The goal of training the RA is to obtain a policy $\pi(\textit{s},\mathcal{I})$ that chooses an action $a$ (an impression list of items) from the set $\mathcal{I}$ given the current state $s$ such that the long-term expected reward (e.g. number of buys) is maximized.

We consider the scenario where a single item\footnote{This can be  extended and generalized to multiple items in a slate \citep{ie2019reinforcement,chen2018generative}.} $i_t \in \mathcal{I}$ is recommended to the user at time $t$, and the response/choice of the user $c_t$ is available to the RA, where the choice is made from a pre-defined set of user-choices such as click, skip, etc.
The immediate reward $r_t$ depends on the choice $c_t$. 
In addition, we consider a ``target choice'', maximizing the frequency of which maximizes the returns, e.g. click-through rate.
For example, if target choice is click, then rewards of 0 for skip, 1 for click can be considered. Here, skip is considered as a negative interaction whereas click is considered as a positive interaction.
A session till time $t$ can thus be represented as $S_t=\{(i_1,c_1,r_1),\ldots,(i_t,c_t,r_t)\}$.
For computational reasons, the last $L$ positive (non-skip) interactions in a session are used to determine the current state of the agent. 

\subsection{State and Action Embeddings \label{sub:state} }
Typically, the item-catalog size $|\mathcal{I}|$ is large (of the order of thousands, or even millions) resulting in an extremely large action space for the RA. 
Furthermore, the state space consisting of sequence of item interactions grows combinatorially in $|\mathcal{I}|$.
We represent items as trainable vectors or embeddings in a dense $d$-dimensional space such that the embeddings of all the items constitute a lookup matrix $\mathbf{I} \in \mathbb{R}^{|\mathcal{I}| \times d} $, where the $j$-th row of  $\mathbf{I}$ corresponds to item $i_j$ represented as $\mathbf{i}_j \in \mathbb{R}^d $ ($j = 1 \dots |\mathcal{I}|$).
Any action $a \in \mathcal{A}$ corresponds to an item, therefore, the action embedding $\mathbf{a} \in \mathbb{R}^d$.
In practice, we find that initializing the item embeddings, i.e. the matrix $\mathbf{I}$ via pre-training a supervised model for next item prediction to be very useful (refer Section \ref{sec:exp}). 
The previously clicked or interacted items in a session are used to predict the next item using Session-based Recommendation with Graph Neural Networks (SRGNN) \citep{wu2018session}. The item embedding matrix after training SRGNN is used to initialize $\mathbf{I}$. Other alternatives include a simple word2vec-like approach \citep{goldberg2014word2vec,zhao2018deep} where items are analogous to words.

The state $s=\{s^1,s^2,\ldots,s^L\}$ of the agent is obtained from the sequence of $L$ most recent non-skip interactions\footnote{As in other approaches \citep{ie2019reinforcement,chen2018generative,bai2019model}, we update the state of the agent only when the action is a non-skip action.} (e.g. clicked items) in a session $S_t$. The corresponding state embedding $\mathbf{s}$ is obtained from the item embedding vectors $\mathbf{s}^k \in \mathbf{I}$ ($k=1\ldots L$) via a bi-directional gated recurrent units (BiGRU) network \citep{cho2014properties} with parameters $\boldsymbol{\theta}$ to obtain the state embedding $\mathbf{s} = \mathbf{W} \mathbf{h}_L + \mathbf{b}$, where $\mathbf{h}_L=BiGRU(\mathbf{s}^1,\ldots,\mathbf{s}^L;\boldsymbol{\theta})$ is the final hidden state of BiGRU, and $\mathbf{W} \in \mathbb{R}^{d \times d}$ and $\mathbf{b} \in \mathbb{R}^d$ are the parameters of the final linear layer. 
These are eventually used to get the value estimates, as detailed later in Section \ref{ssec:distRL}.

\subsection{Constraining the Action Space}
The standard off-policy DRL algorithms like Double Q-learning (hereafter, referred to as DQN) \citep{van2016deep} assume further interactions with the current policy while training with a history of experiences generated by previous iterations of the policy. 
In other words, the initial batch data $\mathcal{B}$ obtained from a behavior policy can be subsequently updated by gathering more experience by interacting with the environment or a model of the environment.
In contrast, batch RL setup additionally assumes that the data set $\mathcal{B}$ is fixed, and no further interactions with the environment are allowed while training. Due to this fixed and limited batch data $\mathcal{B}$, batch RL is not guaranteed to converge. 
When selecting an action $a$, such that $(s, a, s')$ is distant from data contained in the batch $\mathcal{B}$, the estimate $Q_{\boldsymbol{\theta}'}(s', a' )$ (from the target network with parameters $\boldsymbol{\theta}'$ in Double DQN \citep{van2016deep}) may be arbitrarily erroneous, affecting the learning process.
This \textit{overestimation bias} \citep{fujimoto2018off} resulting from a mismatch in the distribution of data induced by the current policy versus the distribution of data contained in $\mathcal{B}$ implies slower convergence of learning due to difficulty in learning a value function for a policy which selects actions not contained in the batch.

To avoid the overestimation bias, we constrain the action-space of the agent for a state $s$ such that it only chooses actions that are likely under the unknown behavior policy $\pi_b$ from which $\mathcal{B}$ is generated, as used in discrete batch-constrained Q-learning (BCQ) \citep{fujimoto2019benchmarking}. 
The action for the next state is selected under the guidance of a state-conditioned generative model $\mathcal{M}$ that approximates the policy $\pi_b$ such that the probability $p_{\mathcal{M}}(a|s)\approx \pi_b(a|s)$. 
Such a behavior cloning neural network is trained in a supervised learning fashion with a cross-entropy loss to solve the $|\mathcal{I}|$-way classification task, $\mathcal{L}_\omega(s,a) = - \text{log}(p_{\mathcal{M}}(a|s))$, over all pairs $(s,a)$ taken from tuples $(s,a,r,s')\in \mathcal{B}$, where $
    p_{\mathcal{M}}(a|s;\omega)=\frac{\text{exp}(\mathbf{s}^T\mathbf{a})}{\sum_{i \in \mathcal{I}}\text{exp}( \mathbf{s}^T\mathbf{i})},
$
$\omega$ being the parameters of the neural network.
The action space of the agent (recommendable items) is restricted to those actions that satisfy $p_\mathcal{M}(a'|s') > \beta$, $\beta\in[0,1)$, as detailed in next subsection.

The training of $\mathcal{M}$ is equivalent to training a deep neural network for SR in a supervised manner, e.g. \citep{liu2018stamp,wu2018session}, where the goal is to predict the next interaction item for a user given past interactions.
The only difference is that while training $\mathcal{M}$, the interactions not only correspond to the positive feedback items but also the skipped items, or items with any response type for that matter.
In this work, we choose SRGNN \citep{wu2018session} (a state-of-the-art graph neural networks based approach for SR) as the neural network architecture for $\mathcal{M}$.

\subsection{Distributional RL Agent\label{ssec:distRL}} 
It has been recently shown that, in practice, better policies can be learned by estimating the value distribution $Z^\pi$ instead of estimating just the expectation of the value $Q^\pi$ \citep{morimura2010nonparametric,bellemare2017distributional,dabney2017distributional}, especially when the environment is stochastic.
Learning the value distribution matters in the presence of approximations which are common in deep RL approaches (e.g. neural networks as value function approximators) \citep{bellemare2017distributional}.
This can be particularly important in the case of SR where the environment is highly stochastic given the variety of users with varying interests and behaviors (refer Section \ref{sec:results}).
BCD4Rec is, therefore, trained in a distributional RL fashion using Implicit Quantile Networks (IQN) \citep{dabney2018implicit}, where $K$ samples from a base distribution, e.g. $\tau \sim U([0,1])$ are reparameterized to $K$ quantile values of a target distribution.
The estimation of action-value for $\tau$-quantile is given by $Q^\tau_{\boldsymbol{\theta}}(s,a) = \mathbf{s}_\tau^T \mathbf{a}$, where $\mathbf{s}_\tau =\mathbf{s} \odot\phi(\tau)$ ($\odot$ is Hadamard product)
for some differentiable function $\phi$ with $\phi: [0,1] \rightarrow R^d$ computing the embedding for the quantile $\tau$.
Note that this form of the value function allows us to efficiently compute the values for all actions (items) in parallel via multiplication of the the item-embedding lookup matrix $\mathbf{I}$ and the vector $\mathbf{s}_\tau$, i.e. using $\mathbf{I}\mathbf{s}_\tau$, indicating the importance of considering latent state and action spaces to handle high dimensional setting like SR. 
The $j$-th dimension of $\phi(\tau)$ is computed as: $\phi_j(\tau):=ReLU(\sum_{i=0}^{n-1}cos(\pi i \tau)w_{ij} + b_j)$
where $w_{ij}$ and $b_j$ for $i={0,\ldots,n-1}$ and $j={0,\ldots,d-1}$ are trainable parameters.

The final loss for training BCD4Rec is computed over all $K^2$ pairs of quantiles based on K estimates each from the current network with parameters $\boldsymbol{\theta}$ and the target network with parameters $\boldsymbol{\theta}'$, and by using $\mathcal{M}$ to constrain the action space as follows:

\begin{equation}\label{eq:loss_bcd}
\begin{aligned}
\mathcal{L}_{BCD}(\boldsymbol{\theta}) &= \frac{1}{K^2}\mathbb{E}_{s,a,r,s'} \left [\sum_\tau \sum_{\tau'} l_\tau \left ( r + \gamma  Q^{\tau'}_{\boldsymbol{\theta}'}(s',a')  - Q^\tau_{\boldsymbol{\theta}}(s,a) \right )\right ],
\\
a' &= \argmax_{ a'|p_\mathcal{M}(a'|s') > \beta} \frac{1}{K}\sum_\tau Q^{\tau}_{\boldsymbol{\theta}}(s',a'),
\end{aligned}
\end{equation}

where, $\tau$ and $\tau'$ are sampled from uniform distribution $U([0,1])$, $l_\tau$ is the quantile Huber loss $l_\tau(\delta) = |\tau - \mathbb{I}({\delta < 0})|L_\kappa(\delta)$ with Huber loss \citep{huber1964robust} $L_\kappa$: $L_\kappa(\delta) = 0.5 \delta^2$ if  $\delta \leq \kappa$, and $\kappa(|\delta| - 0.5\kappa)$ otherwise. An estimate of the value can be recovered through the mean over the quantiles, and the policy $\pi$ is defined by greedy selection over this value: $\pi(s) = \argmax_{a} \frac{1}{K} \sum_\tau Q^\tau_{\boldsymbol{\theta}}(s,a)$.

Refer Appendix \ref{sec:app_a} for a summary of the training procedure for BCD4Rec.

\section{Experimental Evaluation\label{sec:exp}}
We evaluate our approach on two domains:
\textbf{Diginetica (DN)} is a real-world offline dataset from CIKM Cup 2016 data challenge.
We pre-process the data in the same manner as in \citep{wu2018session}. It contains six months of user interaction logs. 
The details related to the items clicked, bought, and skipped in each session are available. 
We use the first 60 days of data to train a user-behavior simulation  environment that we use as a proxy for online testing, and the next two disjoint sets of 30 day data each to i. train the recommender agents, and ii. test the policies in the offline setting. 
During training, rewards considered for different action types are skip:0, click:1, buy:5. 
\textbf{RecSim} \citep{ie2019recsim,ie2019reinforcement} is a recently proposed simulation environment for testing RL agents for SR. 
We consider batch learning on data obtained from three behavioral policies (and refer to these as RecSim-x, x=1,2,3) with different click through rates (CTR) obtained from checkpoints at different iterations while training an IQN agent in online setting using the simulator. 

\textbf{Baselines Considered \label{sec:baselines}}
We consider several non-RL and RL baselines:
i. \textbf{Heuristic-based approaches}: \textit{MostPop}, \textit{SKNN} \citep{jannach2017recurrent} and \textit{STAN} \citep{garg2019sequence} are popular baselines that recommend items in decreasing order of their popularity (MostPop), or based on nearest neighbors defined in terms of similarity of on-going session to sessions in historical logs (SKNN and STAN).  
ii. \textbf{Supervised Deep Neural Networks}: We consider two state-of-the-art SR approaches using supervised learning, namely, \textit{GRU} \citep{hidasi2018recurrent} and \textit{SRGNN} \citep{wu2018session} to
predict the next positive interaction. 

iii. \textbf{Deep RL agents}: We consider double Q-learning \citep{van2016deep} \textit{DQN} and its extension \textit{BCQ} \citep{fujimoto2019benchmarking} that uses batch-constraining (BC). In distributional RL methods, we consider QRDQN \citep{dabney2017distributional}, its batch-constrained version QRBCQ, and IQN \citep{dabney2018implicit}. 
Buy Rate (BR) and Click Through Rate (CTR) are the chosen performance metrics for online evaluation of agents in RecSim and DN, respectively. Coverage@3 (C@3) is used as additional metric to study the diversity of the recommendations made.
Refer Appendix \ref{sec:app_a}, \ref{sec:app_b}, \ref{sec:eval_metrics}, and \ref{sec:app_d} for more details on baselines, datasets, performance metrics, and hyperparameter settings, respectively.

\subsection{Results and Observations\label{sec:results}}

\begin{table*}
    \centering
	\footnotesize
	\caption{Comparison of various approaches with the proposed BCD4Rec on online evaluation metrics BR, CTR and Coverage (C@3). BCD4Rec significantly improves upon the behavior policy, and has performance closest to the best achievable online policy. (Though higher coverage indicates higher diversity in recommendations and less bias, it does not necessarily imply better performance; an exploration policy will have very poor CTR/BR while having very high coverage.) Note: numbers are reported as average over 3 runs.  \label{tab:online}}
	
	\scalebox{0.85}{
	\centering 
		\begin{tabular}{l|c c | c c| c c| c c}
		\hline
		& \multicolumn{2}{|c|}{\textbf{Diginetica (DN)}} & \multicolumn{2}{|c|}{\textbf{RecSim-1}} & \multicolumn{2}{|c|}{\textbf{RecSim-2}} & \multicolumn{2}{|c}{\textbf{RecSim-3}}\\
		\hline	
		\textbf{Method} & \textbf{BR} & \textbf{C@3}  & \textbf{CTR} &\textbf{C@3}  &\textbf{CTR} & \textbf{C@3}  & \textbf{CTR }&\textbf{C@3}  \\
		\hline
		Most pop & 2.0  &\textbf{24.3} &39.5 & 30.0  &41.5 &30.0  &40.3 &30.0 \\
		
		SKNN & 2.0  & 5.3  & 58.1  & 79.0  & 60.2  & 73.5  & 76.1  &66.0  \\

		STAN &2.0  & 6.1  & 61.3  & 79.5  & 66.7 & 75.0   & 77.9  & 67.0 \\
		\hline
		
		GRU &4.2 $\pm$ 0.6 & 11.8 $\pm$ 0.9 &  63.5 $\pm$ 1.7& 75.0 $\pm$ 1.8 &  69.5 $\pm$ 1.1& 77.3 $\pm$ 2.4  & 74.1 $\pm$ 0.9 &68.5 $\pm$ 2.3\\
		
		SRGNN & 3.1 $\pm$ 0.9 &11.1 $\pm$ 0.2 &63.1 $\pm$ 1.3 &75.5 $\pm$ 4.4 &67.8 $\pm$ 2.0&77.2 $\pm$ 2.3  & 77.7 $\pm$ 2.0& 75.0  $\pm$ 2.8\\
		\hline
		
		DQN & 4.8 $\pm$ 0.4 &7.8 $\pm$ 0.5 &50.7 $\pm$ 4.0 &66.2 $\pm$ 2.3 &52.1 $\pm$ 3.1 &65.8 $\pm$ 9.5 & 54.0 $\pm$ 2.7 & 77.5 $\pm$ 3.3\\
		
		BCQ &13.3 $\pm$ 1.1 &10.7 $\pm$ 0.5 &65.9 $\pm$ 1.5& 74.5 $\pm$ 3.6 & 72.3 $\pm 2.3 $ &71.5 $\pm$ 0.4  &76.3 $\pm$ 1.1 &74.9 $\pm$ 2.0\\
		
		QRDQN &14.9 $\pm$ 1.3 &\underline{16.7} $\pm$ \underline{1.2} & 62.9 $\pm$ 1.8 &74.3 $\pm$ 4.3 & 63.9 $\pm$ 1.5 & 73.2 $\pm$ 3.6 &78.2 $\pm$ 1.3 &68.7 $\pm$ 2.8\\
		
		QRBCQ &15.3 $\pm$ 2.3 &14.9 $\pm$ 0.6 &70.8 $\pm$ 0.9 &74.5 $\pm$ 4.4 &74.7 $\pm$ 0.6 &73.8 $\pm$ 6.8   &81.4 $\pm$ 1.9 &74.8 $\pm$ 2.0 \\
		
		IQN &\underline{23.5} $\pm$ \underline{2.0} &12.4 $\pm$ 1.4 &\underline{73.5} $\pm$ \underline{0.7} &\underline{79.5}  $\pm$ \underline{4.0} &\underline{75.9} $\pm$ \underline{2.5} &\textbf{79.8 $\pm$ 2.8} &\underline{81.5} $\pm$ \underline{2.1} &\underline{78.3} $\pm$ \underline{2.8} \\
		
		BCD4Rec &\textbf{24.1} $\pm$ \textbf{1.7} &14.2 $\pm$ 0.3 &\textbf{76.4 $\pm$ 0.8} &\textbf{79.7 $\pm$ 0.8} & \textbf{79.3 $\pm$ 1.2} &\underline{77.8} $\pm$ \underline{4.3} &\textbf{83.2 $\pm$ 1.7} &\textbf{82.1 $\pm$ 1.8}\\
		\hline
		
		\textit{Behavior} & \textit{4.1 } &\textit{23.0}  &\textit{63.1} &\textit{97.0}  &\textit{68.3 } &\textit{90.0 }   &\textit{79.9} &\textit{84.0} \\
		
		\textit{Online} & \textit{26.6}& \textit{21.1 }& \textit{85.9 }&\textit{100.0}  & \textit{85.9}&\textit{100.0}  &\textit{85.9} &\textit{100.0}  \\
		
		\hline
	\end{tabular}}
\end{table*}

\begin{figure*}
	\centering
	\subfigure[\scriptsize Category Prediction Acc. (DN)]{\includegraphics[width=0.32\textwidth]{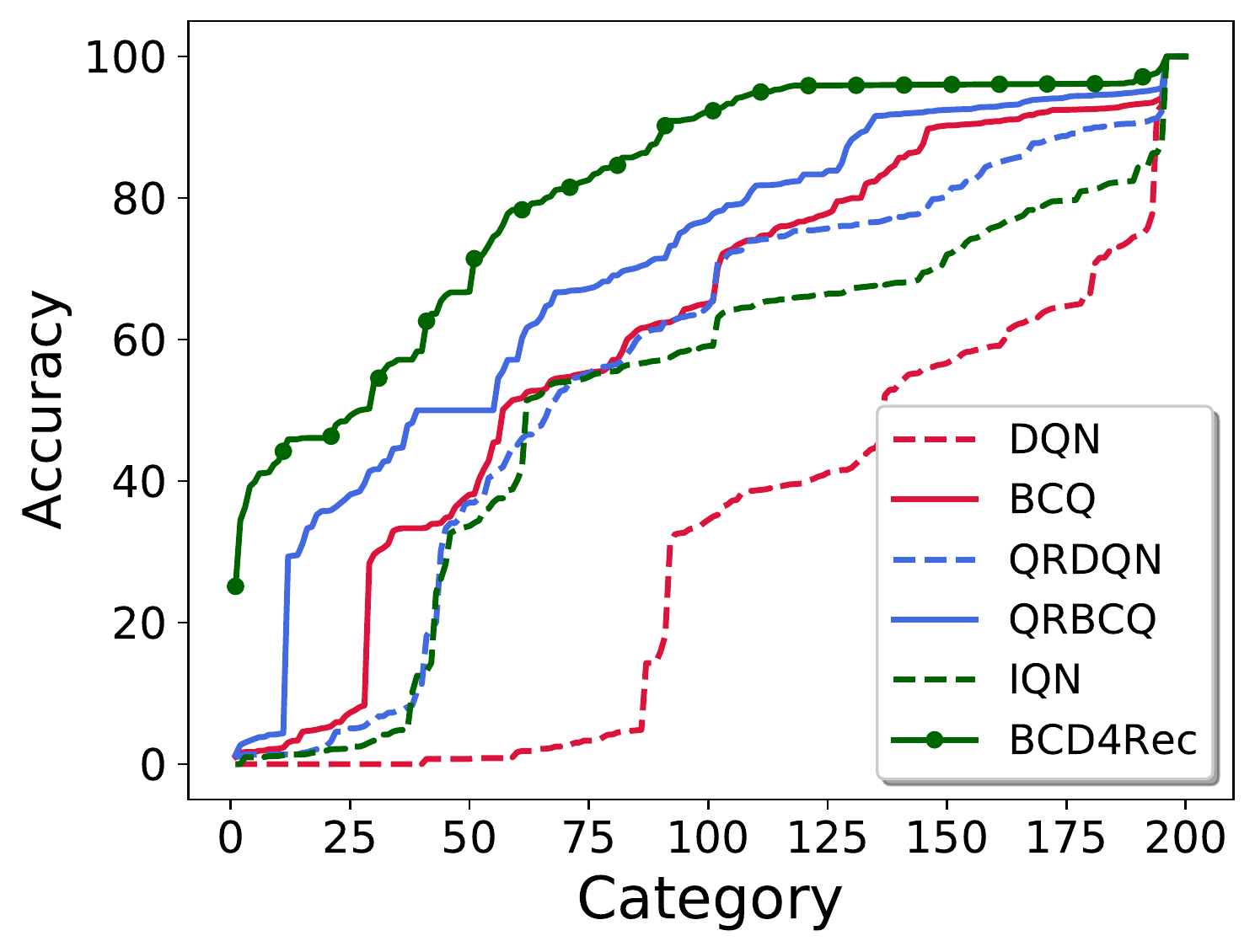}}
	\subfigure[\scriptsize Pop. Bias (DN)]{\includegraphics[width=0.32\textwidth]{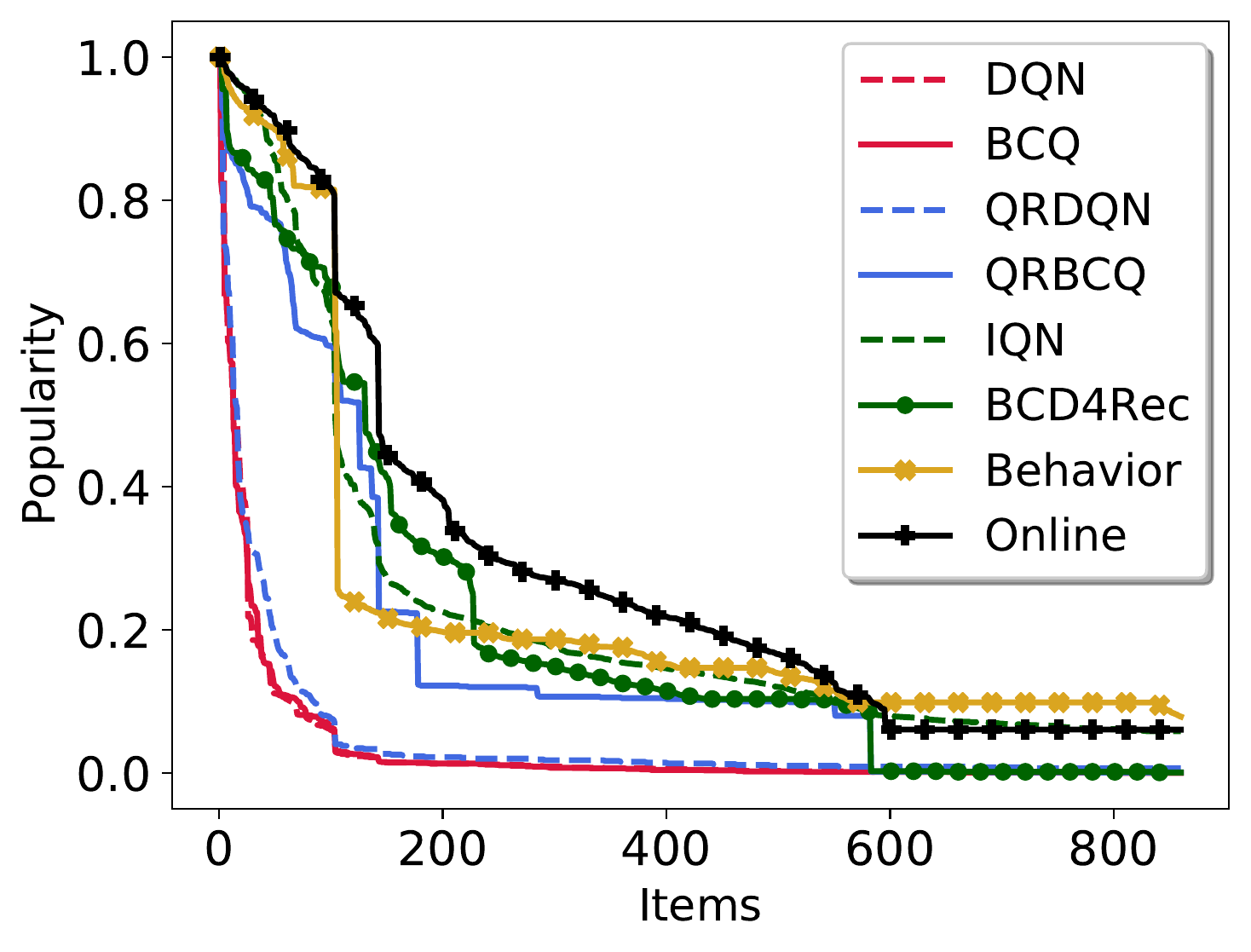}}
	\subfigure[\scriptsize Pop. Bias (RecSim-2)]{\includegraphics[width=0.32\textwidth]{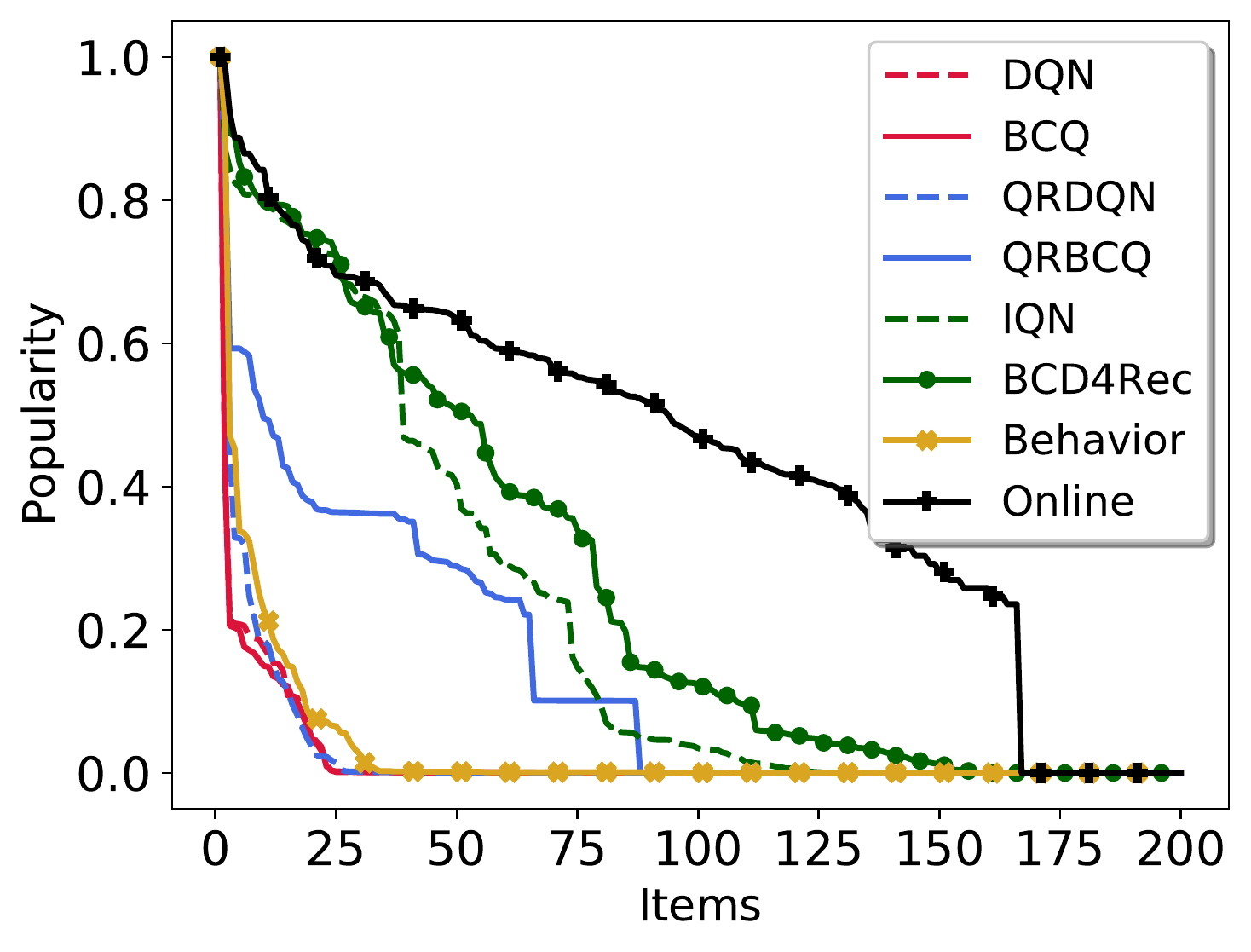}}
	\caption{\textbf{(a)} Using category of the first item in a session (episode) as the target, we evaluate the percentage accuracy of recommending the items from the same category throughout the session. BCD4Rec almost perfectly learns to recommend items from the (latent) category of interest, indicating better handling of  overestimation bias. (We depict the most popular 200 categories out of 825 for clarity.)
	\textbf{(b),(c)} Popularity Bias\label{fig:extrapolation}: BCD4Rec learns to recommend a diverse set of items and has item recommendation distribution close to that of the online policy. DQN, BCQ and QRDQN show very high popularity bias, and in turn, poor diversity in recommended items for DN where the action space is large. (For DN, we depict the most popular 850 items out of $\approx$6.7k for clarity.)}
\end{figure*}

1. From Table \ref{tab:online}, we observe that across all tasks, DQN and all non-RL baselines (heuristics and supervised learning methods) struggle to improve upon the behavior policy.  
This observation is inline with the results from other domains where DQN is shown to suffer in the ``truly'' off-policy batch-constrained learning setting \citep{fujimoto2018off,fujimoto2019benchmarking}, and even the heuristics or supervised methods can mimic the behavior policy at best.
Importantly, \textit{BCD4Rec significantly improves upon the behavior policy in all cases}, substantially bridging the gap between behavior policy and online (best possible) results.
By estimating the return distribution rather than the expected return, the distributional RL methods (QRDQN and IQN) improve upon DQN even without BC. (Refer Appendix \ref{sec:app_e} for more insights and results on ability of various approaches to learn the return distribution). 
This shows the efficacy of distributional RL methods in the previously unexplored batch learning for SR settings in the literature.

2. We observe that BC improves the performance of all RL methods including DQN, QRDQN and IQN. The gains from batch-constraining are maximum for  BCQ vs DQN, while least for the distributional RL method BCD4Rec vs IQN, indicating that BC is critical when using non-distributional RL methods.
Overall, our results indicate that it is \textit{possible to improve upon the behavior policy without having further access to the costly interactions in the real environment} even for the challenging large state and action space settings of SR. 

3. Apart from the evaluation metrics, we study the following property of the learned agents as a proxy to evaluate overestimation bias: 
Each item in DN dataset has a category associated with it (total 825 categories). In each session, the user interacts with items from only one category. Given one or more positive interactions in a session, a good RA should learn to discard items from irrelevant categories by assigning low value estimates to them.
This is a challenging enough problem as the action space is large ($\approx$6.7k items), and erroneously high value estimates even for one irrelevant item from a different category can lead to significant overestimation bias, due to subsequent build-up of error in the absence of further corrections feasible only via further exploration in online setting. 
Figure \ref{fig:extrapolation}(a) demonstrates that DQN, QRDQN and IQN agents are more prone to recommending items from non-relevant categories, and are, in turn overestimating the values for irrelevant actions (items from wrong categories). In contrast, their counterparts using constrained action spaces (BC) during training can additionally rely on $\mathcal{M}$ during training to \textit{guide the agents to learn correct value estimates for the relevant actions by enforcing constraints on the action space, in turn tackling overestimation bias}. 

4. From Figs. \ref{fig:extrapolation} (b) and (c), we observe that distributional RL-based agents not only improve the BR and CTR (i.e. make relevant recommendations), but also reduce the skewness (popularity bias) in the distribution of recommended items, and thus depicting more diversity in the recommended items.
DQN, BCQ and QRDQN depict high popularity bias in RecSim as well as DN. In DN, where action space is large, DQN, BCQ and QRDQN have significantly higher bias (more skew) than even the behavior policy.
Distributional RL methods BCD4Rec, IQN and QRBCQ have low popularity bias which is close to that of the online (best achievable) agent.

\begin{figure*}
	\centering
	\subfigure[\scriptsize DN]{\includegraphics[width=0.23\textwidth]{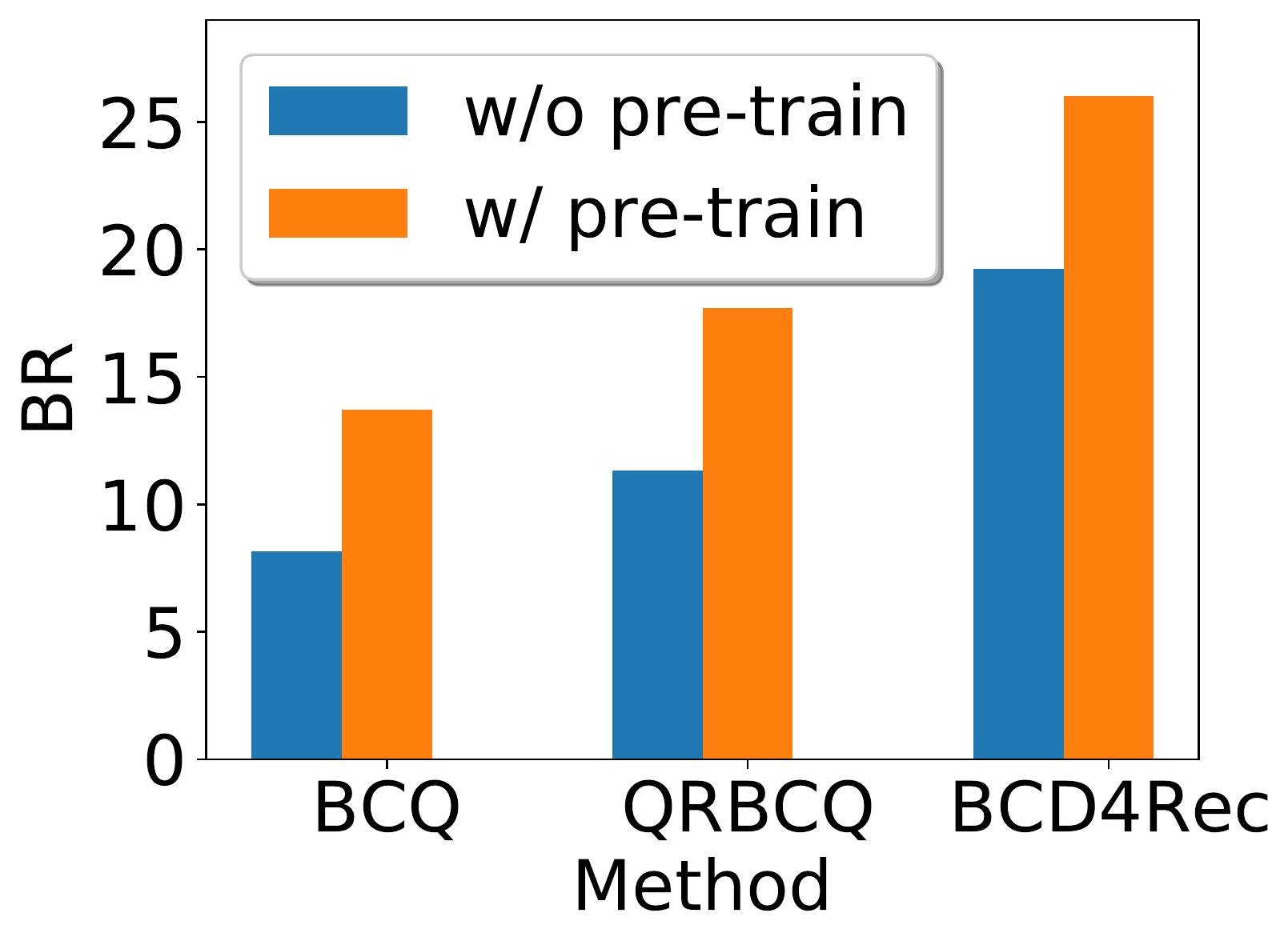}}
	\subfigure[\scriptsize RecSim-2]{\includegraphics[width=0.23\textwidth]{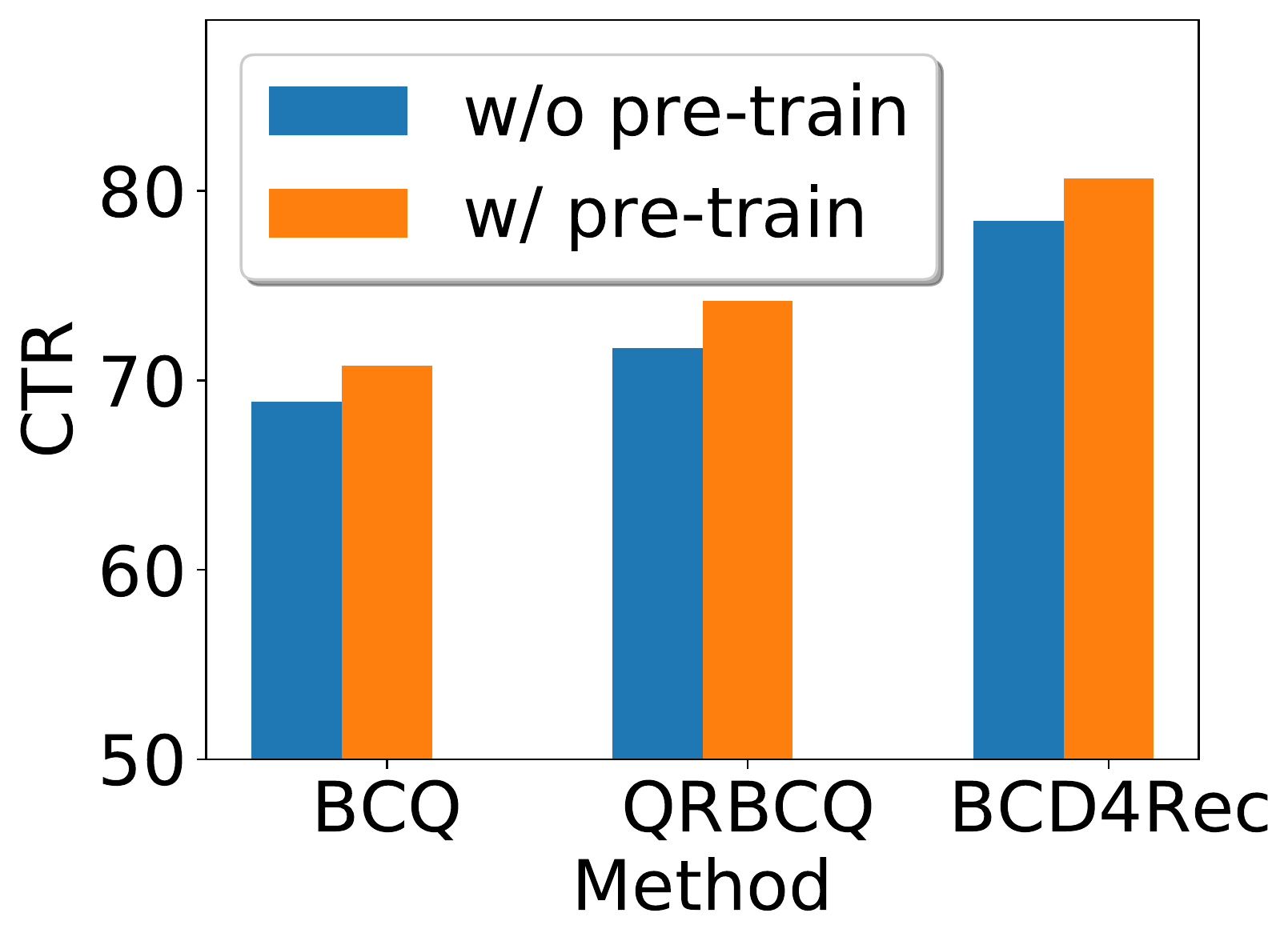}}
	\subfigure[\scriptsize DN]{\includegraphics[width=0.22\textwidth]{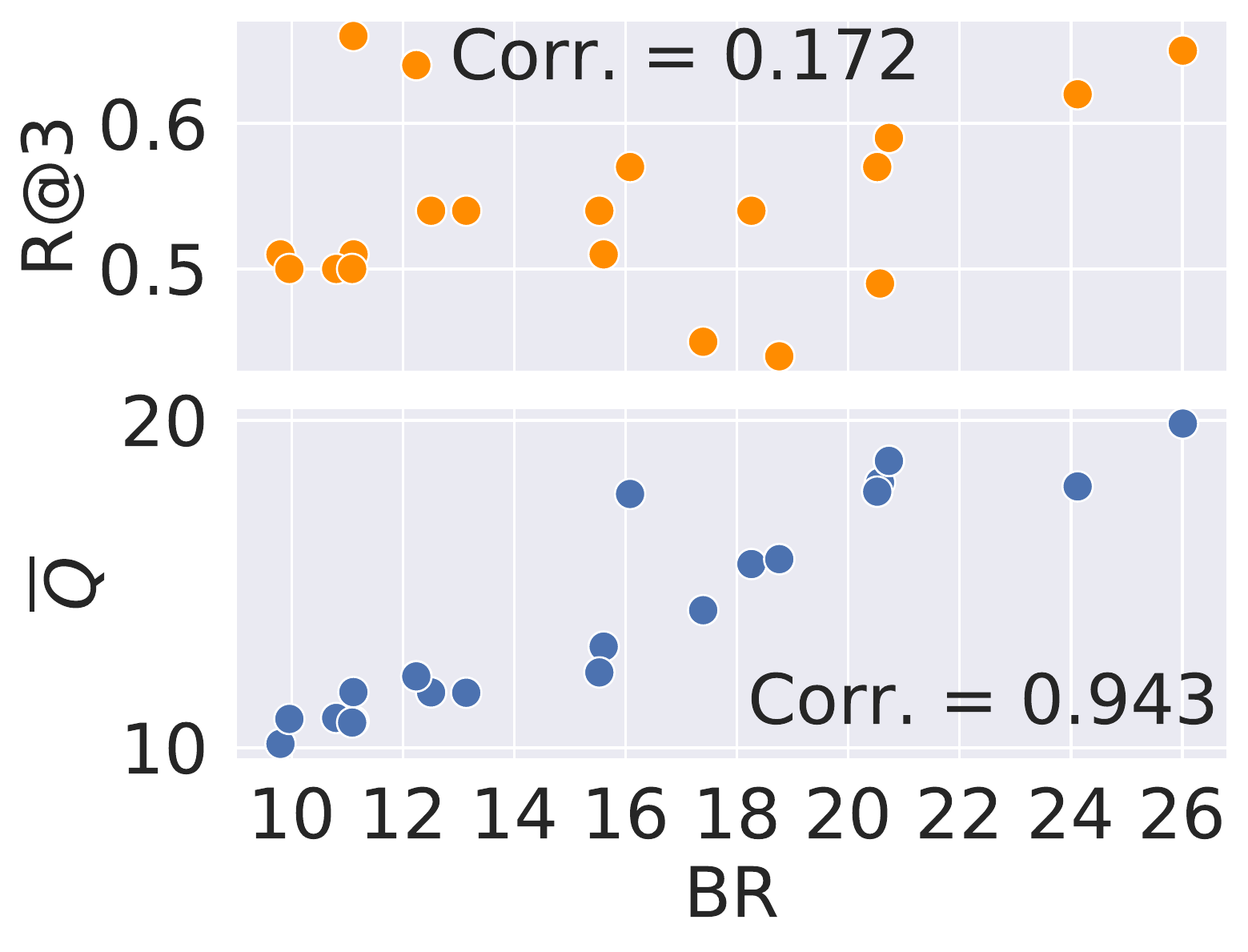}}
	\subfigure[\scriptsize RecSim-2]{\includegraphics[width=0.23\textwidth]{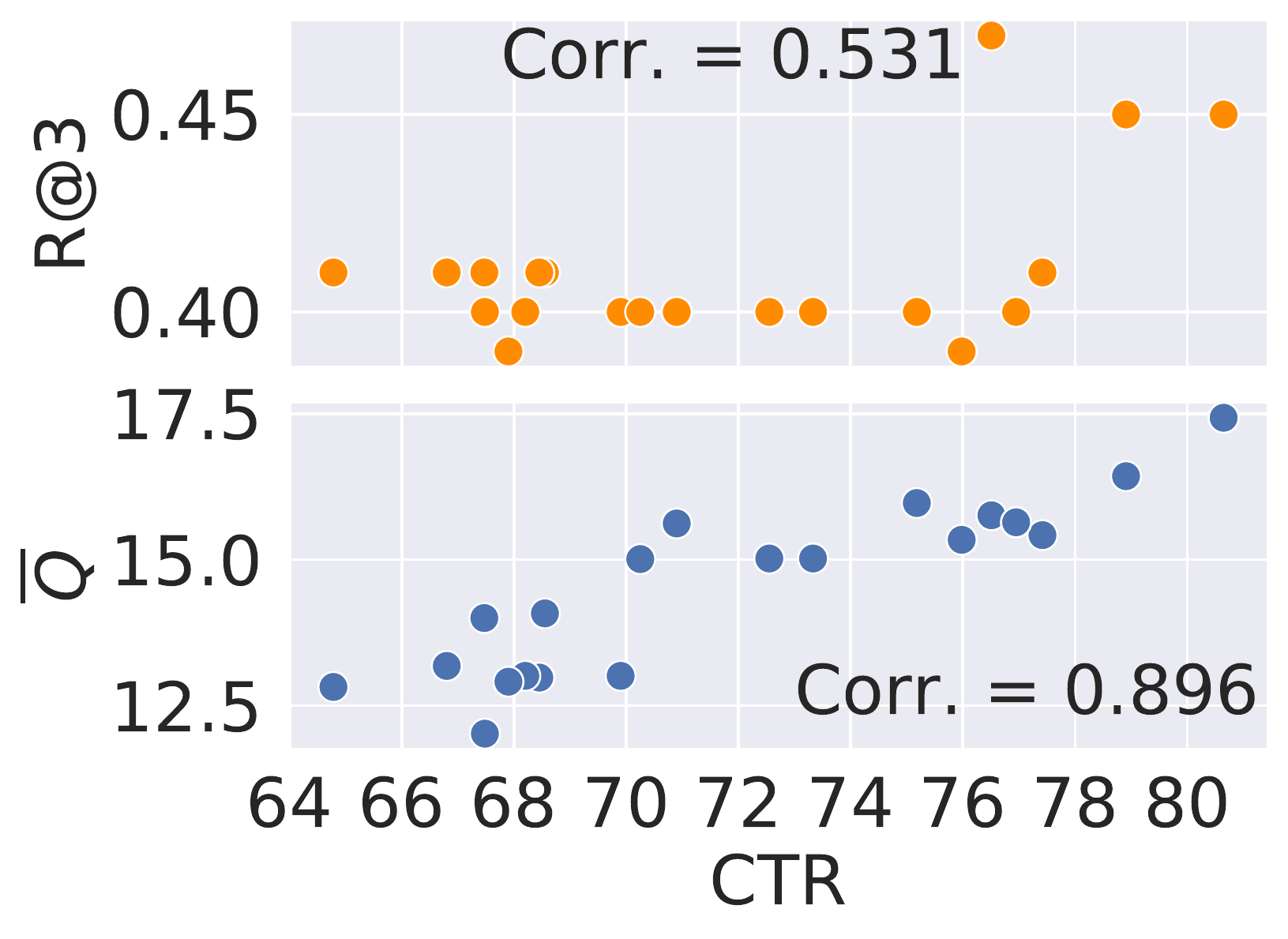}}
	\caption{\textbf{(a),(b)} Pre-training of item embeddings leads to significant gains in BR and CTR. Gains are higher for DN where the action space is larger. \textbf{(c),(d)} Scatter plots depicting correlation between online (BR / CTR) and offline (R@3 / $\Bar{Q}$) evaluation metrics for hyperparameter selection. We found average Q-value estimate $\Bar{Q}$ to be a more reliable metric than the commonly used Recall ($R@3$). \label{fig:thresholding}}
\end{figure*}

5. Pre-training item embeddings using SRGNN on the positive interaction data from offline logs improves the performance of all the agents. The gains are higher in DN where the action space is larger ($\approx$6.7k) compared to RecSim (200) suggesting the importance of pre-training the item embeddings when dealing with large action spaces, as shown in Figs. \ref{fig:thresholding} (a) and (b).

6. Hyper-parameter selection in offline manner is an unsolved problem in batch RL literature \citep{paine2020hyperparameter}.
We find the commonly used Recall metric \citep{bai2019model} is not reliable for hyperparameter selection.  On the other hand, the average Q-value on a hold-out validation set ($\Bar{Q}$) from $\mathcal{B}$ correlates better with the online performance metrics (refer Appendix \ref{sec:eval_metrics} for detailed explanation of metrics). For varying hyperparameter values (for BC threshold $\beta$, number of cosines $n$ and number of quantiles $K$), we compare the performance of BCD4Rec in terms of both offline and online performance metrics, and observe that $\Bar{Q}$ is strongly correlated with online evaluation metrics CTR and BR in comparison to $R@3$, as shown in Figs. \ref{fig:thresholding} (c) and (d). 
We have similar observations for BCQ and QRBCQ agents (refer Appendix \ref{sec:app_d}).

\section{Conclusion and Future Work\label{sec:discussion}}
In this paper, we have studied the problem of batch reinforcement learning (RL) for session-based recommender systems (SR). 
Building upon the recent advances in distributional RL and batch RL, we have proposed a robust approach for batch-constrained distributional RL for SR that does not require exploration.
We have demonstrated the efficacy of the proposed approach on a publicly available simulation environment and a real-world dataset. Our results suggest that: i.  distributional RL and batch-constraining show significant improvements over vanilla Q-learning in the SR setting, ii.  
distributional RL is critical to overcome the popularity bias in the offline logs, iii. pre-training of item embeddings significantly improves the performance in batch RL setting when the action space is large (of the order of 1000s), iv. the commonly used Recall metric is not reliable for hyperparameter selection and the average Q-value on a hold-out validation set from the offline logs correlates better with the online performance metrics.
In future, it will be interesting to i. explore model-based batch RL approaches for SR using recent advances, e.g. \citep{bai2019model,chen2018generative}, ii. evaluate on even larger action spaces, iii. explore ideas at the intersection of causal RL and batch RL \citep{bannon2020causality}, and evaluate the efficacy of other recent advances in batch RL, e.g. \citep{kumar2020conservative,wang2020critic,kidambi2020morel}.

\bibliographystyle{abbrvnat}
\bibliography{rl,gnn-plus,sigir}

\clearpage

\appendix

\section{Details of Deep RL agents used \label{sec:app_a}}
We elaborate on the features contrasting all the RL agents (RAs) as summarized in Table \ref{tab:mcomp}.

\begin{table}[t]
    \centering
  	\caption{Variants of DRL Agents used\label{tab:mcomp}}
  	
 \scalebox{0.8}{\begin{tabular}{l|c|c}
 			\hline			
 			\textbf{Methods} & \textbf{Batch-Constrained}& \textbf{Distributional} \\
 			\hline			
 			DQN &\ding{53}&\ding{53}\\
 			QRDQN &\ding{53}&\checkmark\\
 			IQN &\ding{53}&\checkmark\\
 			BCQ &\checkmark&\ding{53}\\
 			QRBCQ &\checkmark&\checkmark\\
 			BCD4Rec&\checkmark&\checkmark\\
 			
 			\hline
 		\end{tabular}}   
 \end{table}

\subsection{Deep Q-Networks (DQN) \label{ssec:dqn_exp}}
Deep Q-network (DQN) \citep{mnih2015human}, parameterized by $\boldsymbol{\theta}$ is used as a function approximator to estimate the action-value function, i.e., $Q(s, a) \approx Q_{\boldsymbol{\theta}}(s,a)$, while encoding states and actions in terms of real-valued embedding vectors.
We use double DQN \citep{van2016deep} (hereafter, DQN refers to the double DQN variant), which uses two networks $Q_{\boldsymbol{\theta}}$ and $Q_{\boldsymbol{\theta}'}$ to mitigate the overestimation bias of DQN by iteratively minimizing the following  loss $\mathcal{L}_{DQN}(\boldsymbol{\theta})$ estimated over mini-batches of transitions $(s, a, r, s')$ sampled from batch data $\mathcal{B}$ \citep{lin1992self}:

\begin{equation}\label{eq:l}
\begin{split}
\mathcal{L}_{DQN}(\boldsymbol{\theta}) = \mathbb{E}_{s,a,r,s'}[L_\kappa( r + \gamma \max_{a'} Q_{\boldsymbol{\theta}'}(s',a') -  Q_{\boldsymbol{\theta}}(s,a))],
\end{split}
\end{equation}

where $L_\kappa$ is the Huber loss \citep{huber1964robust}: $L_\kappa(\delta) = 0.5 \delta^2$ if  $\delta \leq \kappa$, and $\kappa(|\delta| - 0.5\kappa)$ otherwise; $Q_{\boldsymbol{\theta}'}$ is the target network with parameters $\boldsymbol{\theta}'$ fixed over multiple training steps or update iterations for $\boldsymbol{\theta}$, and $\boldsymbol{\theta}'$ is updated to $\boldsymbol{\theta}$ after a set number of training steps.

\subsection{Distributional RL with Quantile Regression DQN (QRDQN)}
In QRDQN \citep{dabney2017distributional},  a set of $K$ $\tau$-quantiles of the value distribution, $\{\tau_i\}^K = \{\frac{i + 0.5}{K}\}^{K-1}_{i=0}$ is estimated. Instead of estimating just the (expected) value for an action, a $K$-dimensional vector representing the $K$ $\tau$-quantiles is produced. 
So, the overall output of QR-DQN is of size $|\mathcal{A}| \times K$  instead of $|\mathcal{A}|$. 
The loss is computed over all pairs of quantiles as follows: 

\begin{equation}\label{eq:l_drl}
\begin{aligned}
\mathcal{L}_{QRDQN}(\boldsymbol{\theta}) &= \frac{1}{K^2}\mathbb{E}_{s,a,r,s'} \left [\sum_\tau \sum_{\tau'} l_\tau \left ( r + \gamma \max_{a'} Q^{\tau'}_{\boldsymbol{\theta}'}(s',a') - Q^\tau_{\boldsymbol{\theta}}(s,a) \right )\right ],
\end{aligned}
\end{equation}
where $l_\tau$ is the quantile Huber loss $l_\tau(\delta) = |\tau - \mathbb{I}({\delta < 0})|L_\kappa(\delta)$.
An estimate of the value can be recovered through the mean over the quantiles, and the policy $\pi$ is defined by greedy selection over this value: $\pi(s) = \argmax_{a} \frac{1}{K} \sum_\tau Q^\tau_{\boldsymbol{\theta}}(s,a)$.

\begin{figure} 
	\centering
	\subfigure[Proposed Approach BCD4Rec]{\includegraphics[scale=0.25, trim=0cm 0cm 0cm 4cm, clip=true]{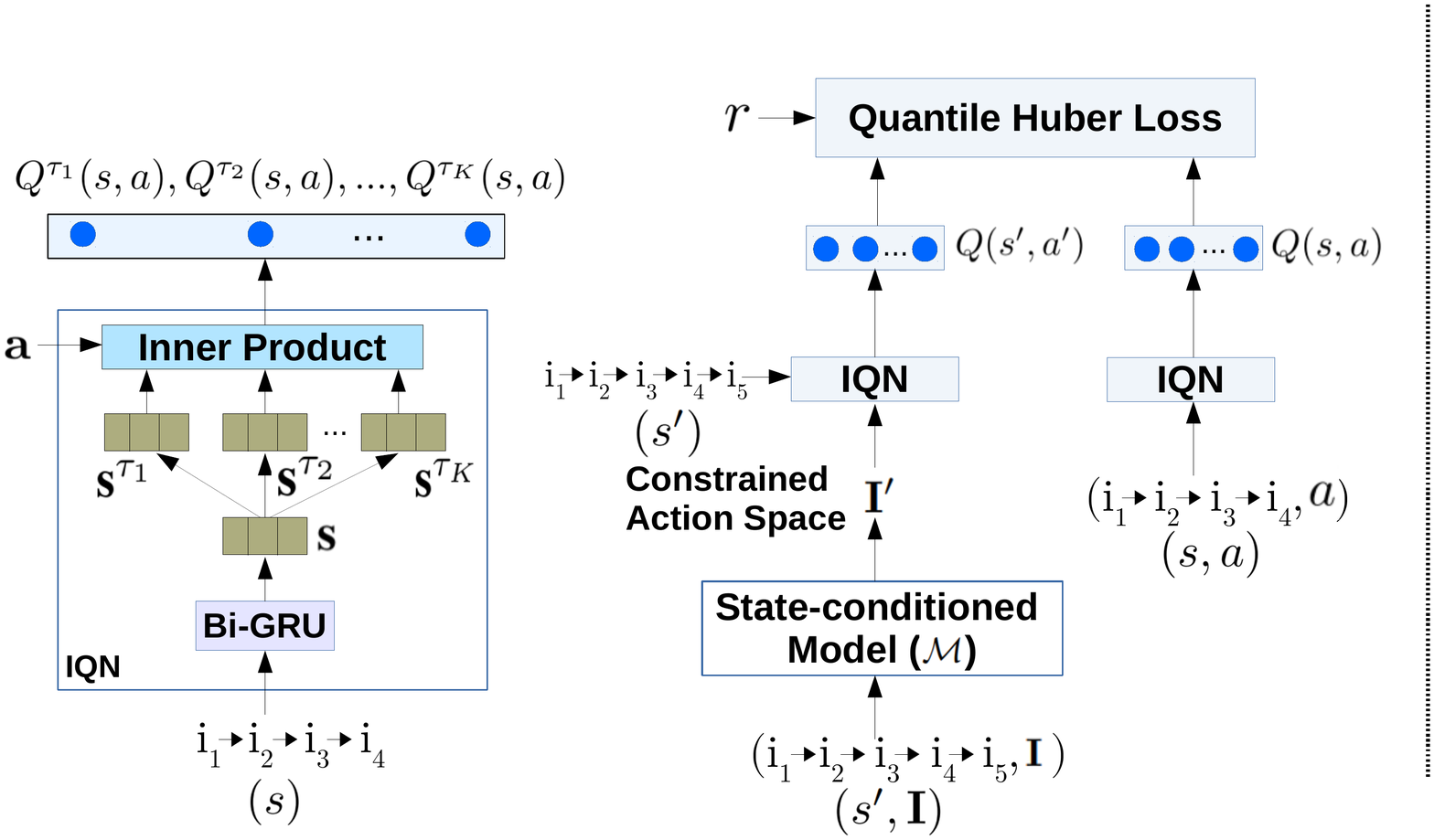}}
	\subfigure[Vanilla DQN]{\includegraphics[scale=0.25, trim=3cm 1.5cm 3cm 4cm, clip=true]{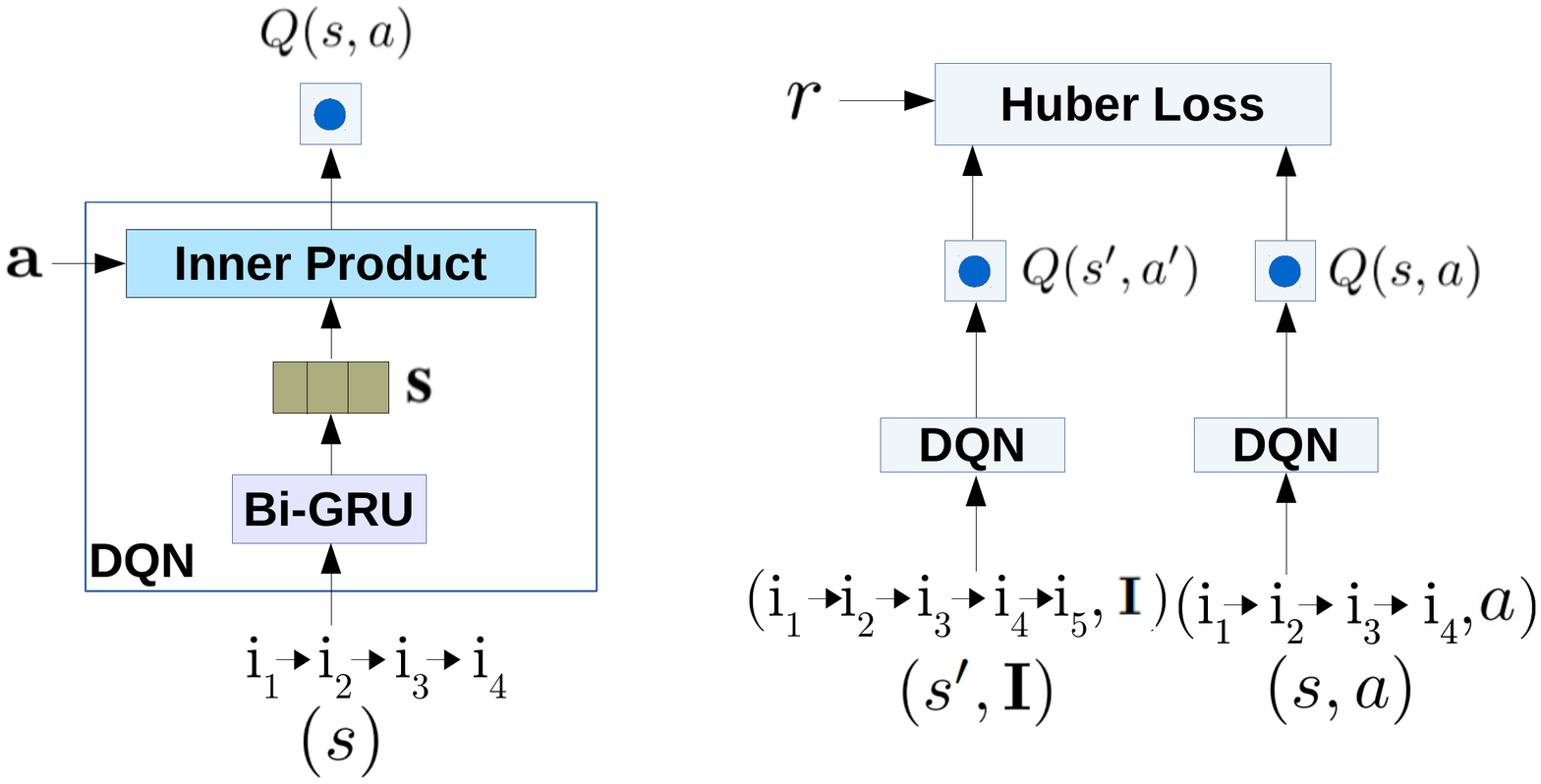}}
	\caption{Handling of a tuple $(s,a,r,s')$ in the proposed approach (BCD4Rec) in contrast to a traditional DQN. The IQN module in BCD4Rec estimates $K$ quantiles of the value distribution while vanilla DQN only estimates the expected (mean) value. Furthermore, the state-conditioned model $\mathcal{M}$ restricts the action space for the BCD4Rec agent.
	While italic $s$ and $a$ denote the state and action, bold $\mathbf{s}$ and $\mathbf{a}$ denote the state and action embeddings, respectively.
	In the example considered, we assume the action $a$ to consist of item $i_5$ that is clicked by the user, leading to updated state $s'=(i_1,i_2,i_3,i_4,i_5)$.\label{fig:app}}
\end{figure}

The batch-constrained variants of DQN and QRDQN, i.e. \textbf{BCQ} and \textbf{QRBCQ} are also trained using losses $\mathcal{L}_{DQN}$ and $\mathcal{L}_{QRDQN}$ respectively, but with the additional action constraining criteria for $a'$:
\begin{equation}
    a' = \argmax_{ a'|p_\mathcal{M}(a'|s') > \beta} \frac{1}{K}\sum_\tau Q^{\tau}_{\boldsymbol{\theta}}(s',a'),
\end{equation}
which is same as that used for BCD4Rec (refer Equation \ref{eq:loss_bcd}).

\subsection{BCD4Rec}
Here we provide additional details for training BCD4Rec in Algorithm 1, and a schematic of BCD4Rec contrasting it with vanilla DQN in Fig. \ref{fig:app}. \textbf{IQN} without batch constraining is equivalent to BCD4Rec with $\beta=0$. 

\begin{algorithm}[t]\label{algo:bcd4rec}
	\caption{BCD4Rec} 
	\begin{algorithmic}[1]
	    \State Input: Batch $\mathcal{B}$, number of iterations $T$, $targetUpdateRate$, mini-batch size $N$, $\mathcal{I}$.
	    \State Initialize the Q-network $Q^{\tau}_{\boldsymbol{\theta}}$ (initialize  the item embeddings using pre-trained embeddings), conditional model $\mathcal{M}$ and target network $Q^{\tau'}_{\boldsymbol{\theta}'}$ with $\boldsymbol{\theta}' \leftarrow \boldsymbol{\theta}$.
		\For {$t=1,2,\ldots T$}
			\State Sample mini-batch $M$ of $N$ transitions $(s,a,r,s')$ from $\mathcal{B}$.
			\State $a' = \argmax_{ a'|p_\mathcal{M}(a'|s') > \beta} \frac{1}{K}\sum_\tau Q^{\tau}_{\boldsymbol{\theta}}(s',a')$ 
            \State $ \boldsymbol{\theta} \leftarrow \argmin_{\boldsymbol{\theta}} \mathcal{L}_{BCD}(\boldsymbol{\theta})$
            
			\State $\omega \leftarrow \argmin_{\omega} - \sum_{(s,a) \in M} \log p_\mathcal{M}(a|s;\omega)$
			\State If $t$ mod $targetUpdateRate$ = 0 : $\boldsymbol{\theta}' \leftarrow \boldsymbol{\theta}$   
		\EndFor
	\end{algorithmic} 
\end{algorithm}

\section{RecSim Simulation Environment \label{sec:app_b}} 
\begin{algorithm}[t]\label{algo:recsim_algo}
	\caption{RecSim: Interest Evolution Environment} 
	\begin{algorithmic}[1]
	\State Input: no. of categories, $C$
    \State \textbf{User model:}
            
            \State Interest vector of user $u$, $\mathbf{u} = [I_1, I_2, ..., I_C]$, where $I_c \sim U([-1,1])$ and $I_c$ is user's interest in category $c$
            \State \textbf{Item model:}
            
            \State One-hot category vector of item $i$, $\mathbf{i} \in \{0,1\}^{C}$
            \State \textbf{User choice model:}
            \State Given a slate ($sl$) of size, $k+1$ (i.e. list of $k$ recommended items by agent and one skip item),
            \State Position of item $i$ in slate $pos(i) \in \{0,1, ..., k\}$ 
            \State User $u$'s interest for item $i$, 
              
            \State
            $
                I(u,i) = 
                \begin{cases}
                    \mathbf{u}^T\mathbf{i}, 
                    & \text{if } pos(i) \in \{0,1, ..., k-1\}\\
                    s_{u},  & \text{if } pos(i) = k\\
                    
                    \end{cases}
            $
            \State $p(u,i)  = \frac{I(u,i)}{\sum_{j \in sl} I(u,j)}$
            \State \textbf{User interest updation:}
            \State $I_{c}$: user $u$'s interest in category $c$ whose item is consumed
            \State $\Delta(I_{c}) = (-y|I_{c}|+y). (1-I_{c}) $ , $y \in [0,1]$
            \State $I_{c} \leftarrow I_{c} + \Delta(I_{c})$ with probability $[I(u,i) + 1]/2$
            \State $I_{c} \leftarrow I_{c} - \Delta(I_{c})$ with probability $[1 - I(u,i)]/2$
\end{algorithmic} 
\end{algorithm}

We consider the Interest Evolution environment of RecSim\footnote{\url{https://github.com/google-research/recsim}} \citep{ie2019recsim}, where the goal is to evaluate RL algorithms to keep a user engaged for as long as possible (we consider the maximum episode length as 20) by showing relevant items that the user would be interested in.
This environment consists of three main modules: i. \textit{user model}, ii. \textit{item model}, and iii. \textit{user choice model}, as summarized in Algorithm $2$. 
This environment has two user response types: click and skip.
A user $u$ is presented with a slate $sl$ consisting of $k$ items and a special skip item such that the effective slate size is $k+1$. The interest or the relevance score $I(u,i)$ of an item $i$  for the user $u$ is defined as per line $10$ of the algorithm, while $s_u$ corresponds to a score for the skip item (in this work, we use the relevance score for the second-most relevant item for a user as the relevance score for the skip item). 
The relevance scores for the recommended items are used to get the probability of clicking an item from a given slate. 
For each item $i \in sl$, this probability is computed as per line $11$, and the action by the user is drawn as per this probability distribution. 
If $u$ clicks on $i$, the relevance score or the interest for the corresponding category is updated as per lines $13-16$. Lines $15-16$ ensure that $u's$ interests are reinforced as the episode progresses, i.e. if $u$ clicks and consumes an item from a category where she had high interest to begin with, the interest in that category is likely to go up.  
We consider the default settings of this environment with $C=20$ and $y$ as $0.3$. 

We consider a random (exploration) policy as one of the behavior policy which is referred to as RecSim-1, and results in batch data with lowest CTR.
We train an IQN agent (variant of BCD4Rec with $\beta$ = 0) from scratch with $\epsilon$-greedy exploration (where $\epsilon$ degrades linearly as used in \citep{mnih2015human}) in online manner. We train this for $1k$ episodes and select behavior policies at different timesteps of training, referred as RecSim-2 and RecSim-3, respectively. 
We keep the user interest vector $\mathbf{u}$ latent (except for optimal policy) while training RL agents (online/offline) and non-RL baselines to mimic the SR scenario. For online policy, we consider $\mathbf{u}$ to be fully observable to an agent.

The behavior policies are then used to generate logs as batch data for evaluating various approaches in batch RL setting. During training, the default rewards of skip:0 and click:4 are used. 
The agents trained using the batch data are compared on 200 previously unseen new users.

\section{Evaluation Metrics \label{sec:eval_metrics}} 
While all RAs are trained in offline fashion using batch data, we evaluate them under two scenarios: i. online, and ii. offline, as is common in literature \citep{bai2019model,chen2018generative}.
\begin{table}[t]
    \centering
  	\caption{Details of the datasets used. Here: s:skip, c: click, b: buy. \label{tab:datastats}}
  	
 \scalebox{0.91}{\begin{tabular}{l|c|c}
 			\hline			
 			\textbf{Statistics} & \textbf{Diginetica}& \textbf{RecSim} \\
 			\hline			
 			\#train sessions  &4843&2000\\
 			\#train tuples &70k &30k\\
 			\#test sessions &1436&200\\
 			\#items&6666&200\\
 			Response Types&\{s,c,b\}&\{s,c\}\\
 			Target Response&b&c\\
 			\hline
 		\end{tabular}}   
\end{table}
\begin{table}
    \centering
	\footnotesize
	\caption{Hyperparameters considered and the best hyperparameters obtained using $\mathbf{\Bar{Q}}$. \label{tab:hyperparameters}}
	\scalebox{0.86}{
	\centering 
		\begin{tabular}{l |c| c| c c}
		\hline
		\textbf{Hyper-parameter (Algorithm) } & \textbf{Range Tried} & \textbf{Selected-DN} & \textbf{Selected-RecSim}   \\
		\hline
		Quantiles $K$ $(DQN/BCQ)$ &${1}$ &$1$ & $1$ \\
		\hline
		Quantiles $K$ $(QRDQN/QRBCQ)$ &${5,7,10}$ &$5$ & $5$ \\
		\hline
		Quantiles $K$ $(IQN/BCD4Rec)$ & ${5,7,10}$ &$5$ &$10$ \\
		\hline
		Cosines Number $n$ $(IQN/BCD4Rec)$ &${32,64,128}$ & $64$ & $128$\\
		\hline
		BC Threshold $\beta$ $(BCD4Rec/QRBCQ/BCQ)$  &${0.1,0.3,0.5,0.7,0.9}$ & $0.3$ & $0.5$\\
		\hline
		 Learning Rate (All) & ${0.0003,0.001,0.003}$  & $0.003$ & $0.003$ \\
		\hline
		HiddenSize $d$ (All) & ${100}$ & $100$ & $100$ \\
		\hline
		Bi-GRU Layers (All) &${2,3}$ & $2$& $2$ \\ 
		\hline
		Bi-GRU Hidden Units (All) & ${d/2 \times K}$ & ${d/2 \times K}$ & ${d/2 \times K}$ \\
		\hline
		Discounted Factor $\gamma$ (All) &$0.9$ & $0.9$& $0.9$ \\
		\hline
		Optimizer &$ADAM$ & $ADAM$ & $ADAM$ \\
		\hline
		Recent positive interactions $L$ & $10$ & $10$ & $10$ \\
		\hline
		Mini-Batch Size &$64$ &$64$ & $64$  \\
		\hline
	\end{tabular}}

\end{table}

\textbf{Metrics for online testing}: Depending upon the target response type that needs to be maximized by the RA, we compute the \textbf{CTR} (click through rate) or the \textbf{BR} (buy rate) as the percentage of responses corresponding to the target response type (i.e. buy for DN, click for RecSim) across all episodes or sessions. 
\textbf{C@X} (Coverage@X) denotes the percentage of items from $\mathcal{I}$ that are recommended at least once across test episodes by the agent within top-$X$ items at any recommendation step.

\textbf{Metrics for Offline Evaluation}: 
\textbf{i) R@X (Recall@X):} Given the initial interactions from a test session, the task is to re-rank the future interacted ground truth items from the session. We compute the standard $R@X$ metric as the percentage of times the eventually clicked or bought items appear in the top-$X$ items in the re-ranked list \citep{bai2019model}. \textbf{ii) $\mathbf{\Bar{Q}}$:}  Average over the q-values of the evaluation policy for the given state distribution $\mathbb{E}_{s \sim \mathcal{B}} [Q_{\theta}(s,a)]$, i.e. the average $Q$ across all states for the action $a$ chosen as per the RA policy (this is similar in spirit to the $V_0$ metric introduced recently in \citep{paine2020hyperparameter}).

\section{Pre-processing and Hyperparameters Selection \label{sec:app_d} }

\begin{table*}
    \centering
	\footnotesize
	\caption{Pearson correlation coefficient (PCC) between online and offline evaluation metrics for various recommender agents. PCC is computed over the original values for the metrics (value-based), and by ranking the values and computing correlation over the ranks (rank-based). We observe that $\mathbf{\Bar{Q}}$ has higher rank-based as well as value-based correlation with online metrics. \label{tab:pearson}}
	
	\scalebox{0.9}{
	\centering 
		\begin{tabular}{l|c c c c| c c c c}
		\hline
		& \multicolumn{4}{|c|}{\textbf{Diginetica}} & \multicolumn{4}{|c}{\textbf{RecSim-2}} \\
		\hline
		& \multicolumn{2}{|c|}{\textbf{Value-based}} & \multicolumn{2}{|c|}{\textbf{Rank-based}} & \multicolumn{2}{|c|}{\textbf{Value-based}} & \multicolumn{2}{|c}{\textbf{Rank-based}} \\
		\hline	
		\textbf{Algorithm} & \textbf{R@3} & $\mathbf{\Bar{Q}}$ & \textbf{R@3} & $\mathbf{\Bar{Q}}$ & \textbf{R@3} & $\mathbf{\Bar{Q}}$ & \textbf{R@3} & $\mathbf{\Bar{Q}}$ \\
		\hline
		BCQ & 0.784 & \textbf{0.786} & \textbf{0.975} & 0.900 & \textbf{0.705} & 0.633 & \textbf{0.700} & 0.600 \\
		
		QRDQN & 0.564 &\textbf{0.941}  & 0.627 & \textbf{0.958} & 0.316 & \textbf{0.653} & 0.174 & \textbf{0.768} \\
		
		BCD4Rec & 0.172 & \textbf{0.943} & 0.227 & \textbf{0.961} & 0.531 & \textbf{0.896} & 0.204 & \textbf{0.898} \\
		\hline
	\end{tabular}}
	
\end{table*}

\begin{figure*}
	\centering
	\subfigure[\scriptsize BCQ (DN)]{\includegraphics[width=0.24\textwidth]{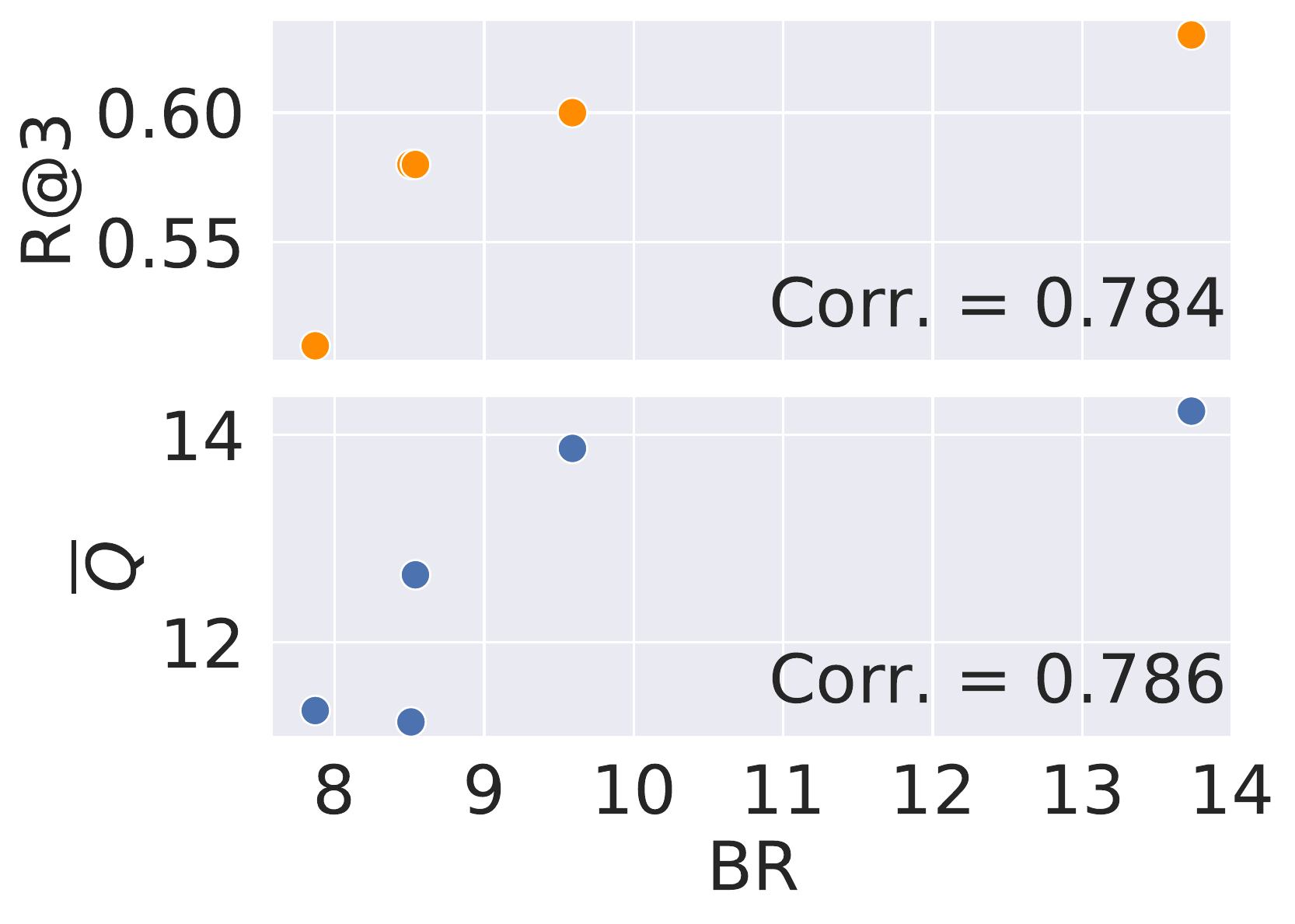}}
	\subfigure[\scriptsize QRBCQ (DN)]{\includegraphics[width=0.23\textwidth]{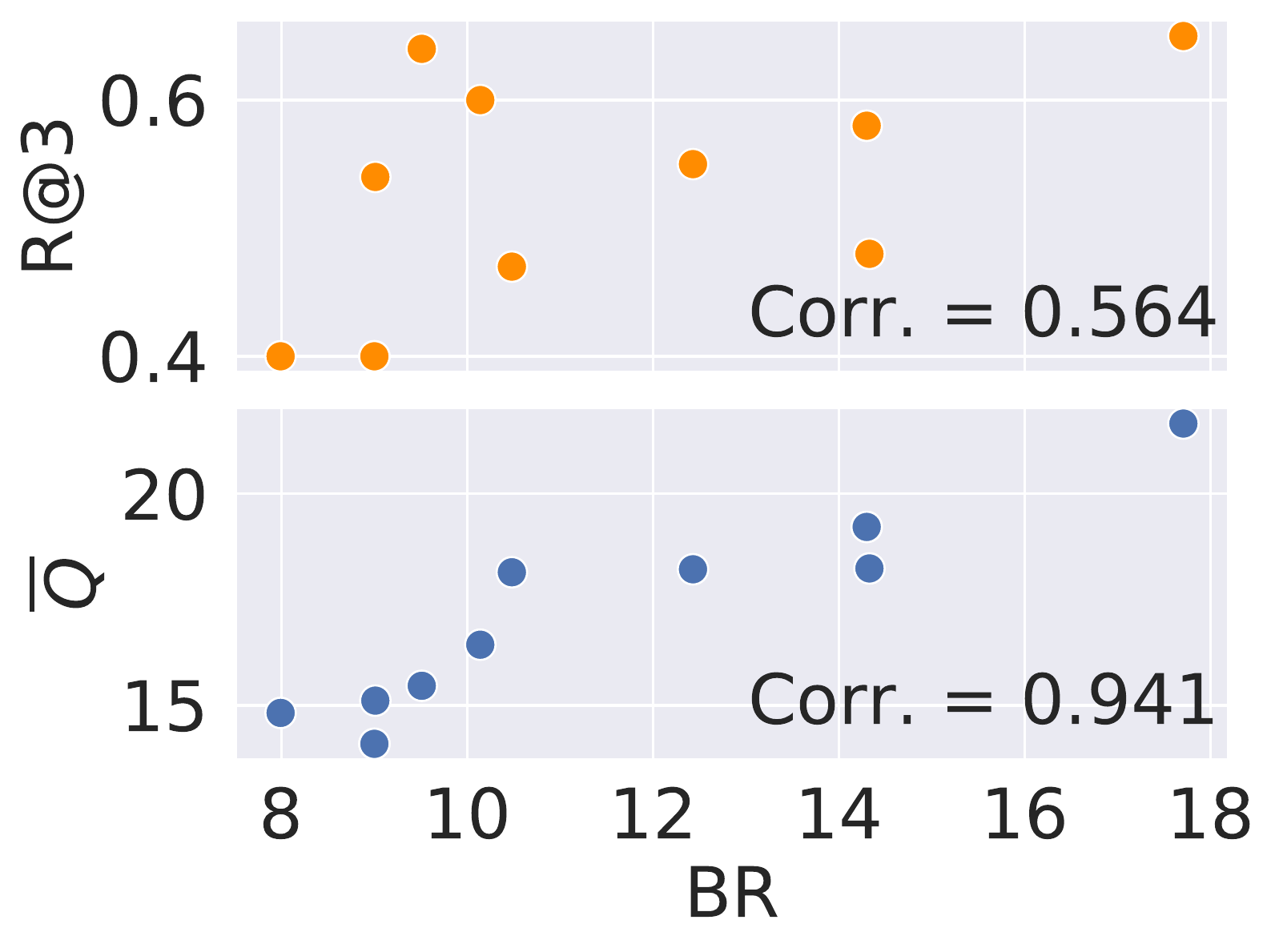}}
	\subfigure[\scriptsize BCQ (RecSim-2)]{\includegraphics[width=0.24\textwidth]{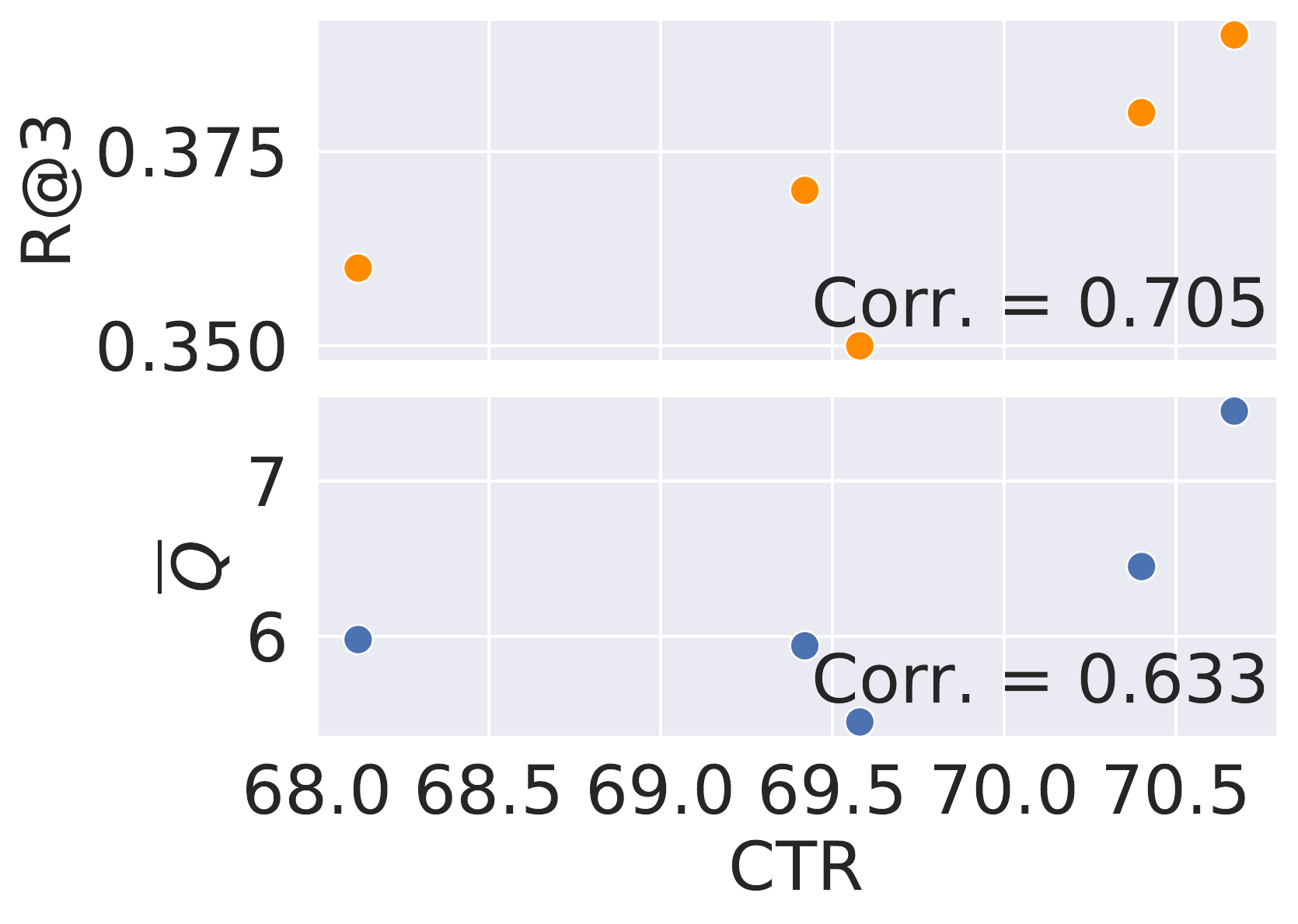}}
	\subfigure[\scriptsize QRBCQ (RecSim-2)]{\includegraphics[width=0.23\textwidth]{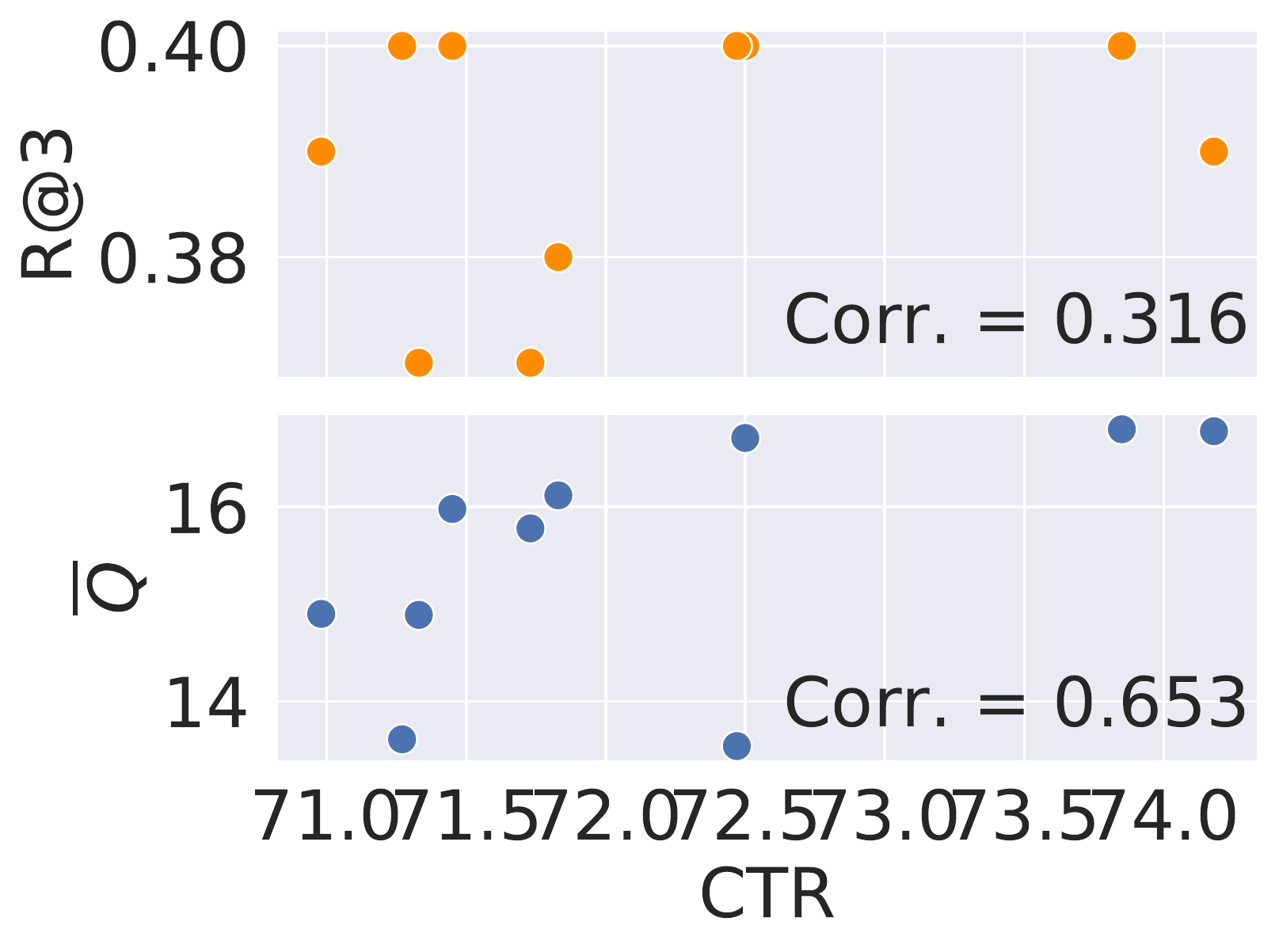}}
	\caption{Comparison of online evaluation metrics (BR/CTR) with offline evaluation metrics $(R@3/\Bar{Q})$ for varying values of $\beta$ and $K$. The higher correlation between online metric BR/CTR and the offline metric $\Bar{Q}$ indicates higher reliability of $\Bar{Q}$ over $R@3$ for hyperparameters selection in batch RL settings. Same is observed for BCD4Rec (refer Section \ref{sec:results}). \label{fig:thresholding_appendix}}
\end{figure*}

\begin{table}
    \centering
	
	\caption{Comparison of the learned value distributions of various recommender agents against the corresponding value distribution from online policy in terms of Wasserstein distance metric for different user types for RecSim-2. \label{tab:wasserstesin}}
	\scalebox{0.85}{
	\centering 
		\begin{tabular}{l|c|c|c|c|c|c}
		\hline
		\textbf{Users} & \textbf{DQN} & \textbf{BCQ} & \textbf{QRDQN} & \textbf{QRBCQ} & \textbf{IQN} & \textbf{BCD4Rec} \\
		\hline
		$User_1$ & 0.055& 0.039& 0.051& 0.047& 0.039& \textbf{0.031} \\
		\hline
		$User_2$ & 0.053 & 0.043&0.047 &0.044 &0.047 &\textbf{0.034} \\
		\hline
		$User_3$ &0.056 &0.052  &0.040 &0.016 &0.036 &\textbf{0.010} \\
		\hline
	\end{tabular}}
	
\end{table}

\begin{figure*}
    \centering
	\subfigure[\scriptsize $user_1$ ($s_0,a_0$)]{\includegraphics[width=0.272\textwidth]{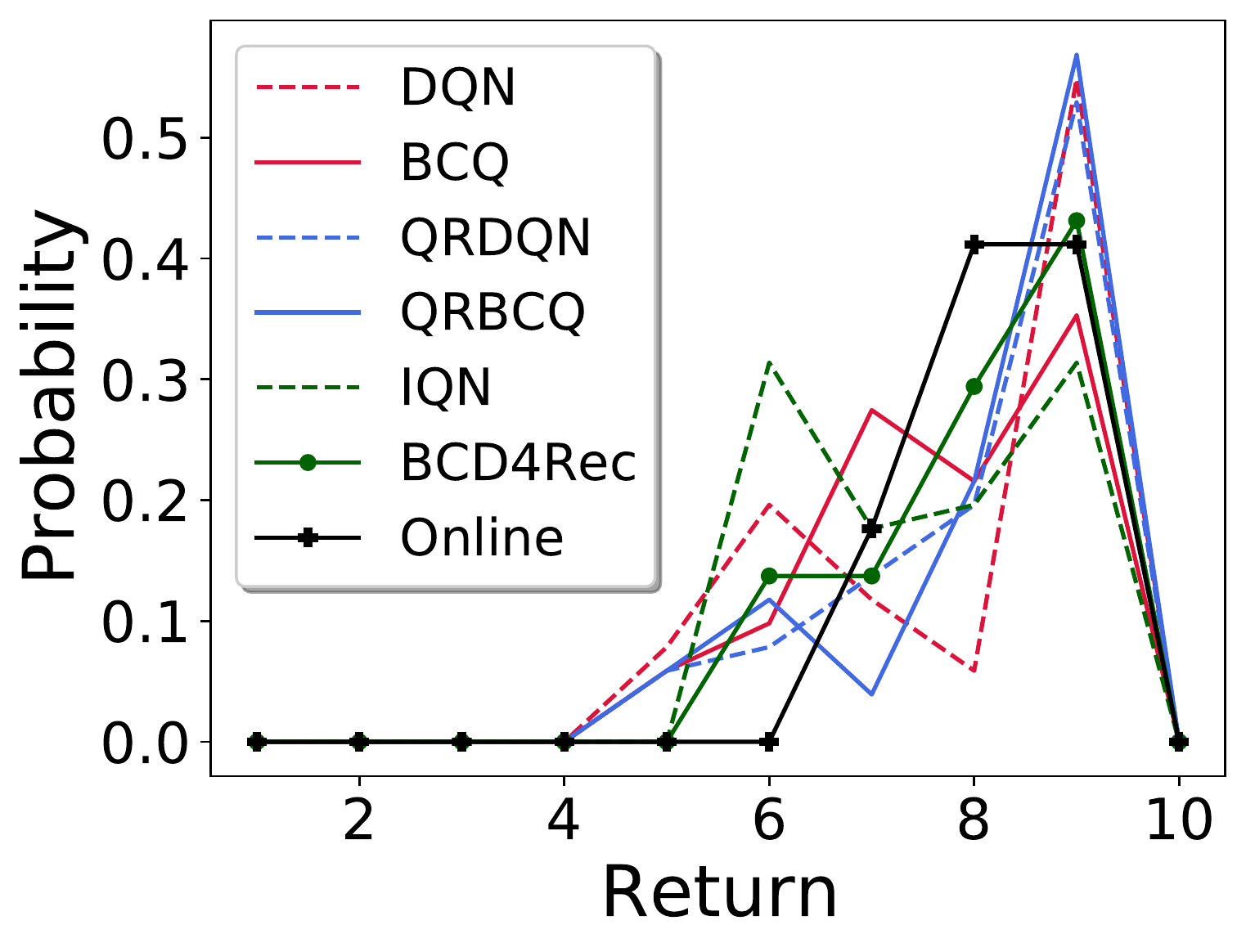}}
	\subfigure[\scriptsize $user_2$ ($s_0,a_0$)]{\includegraphics[width=0.272\textwidth]{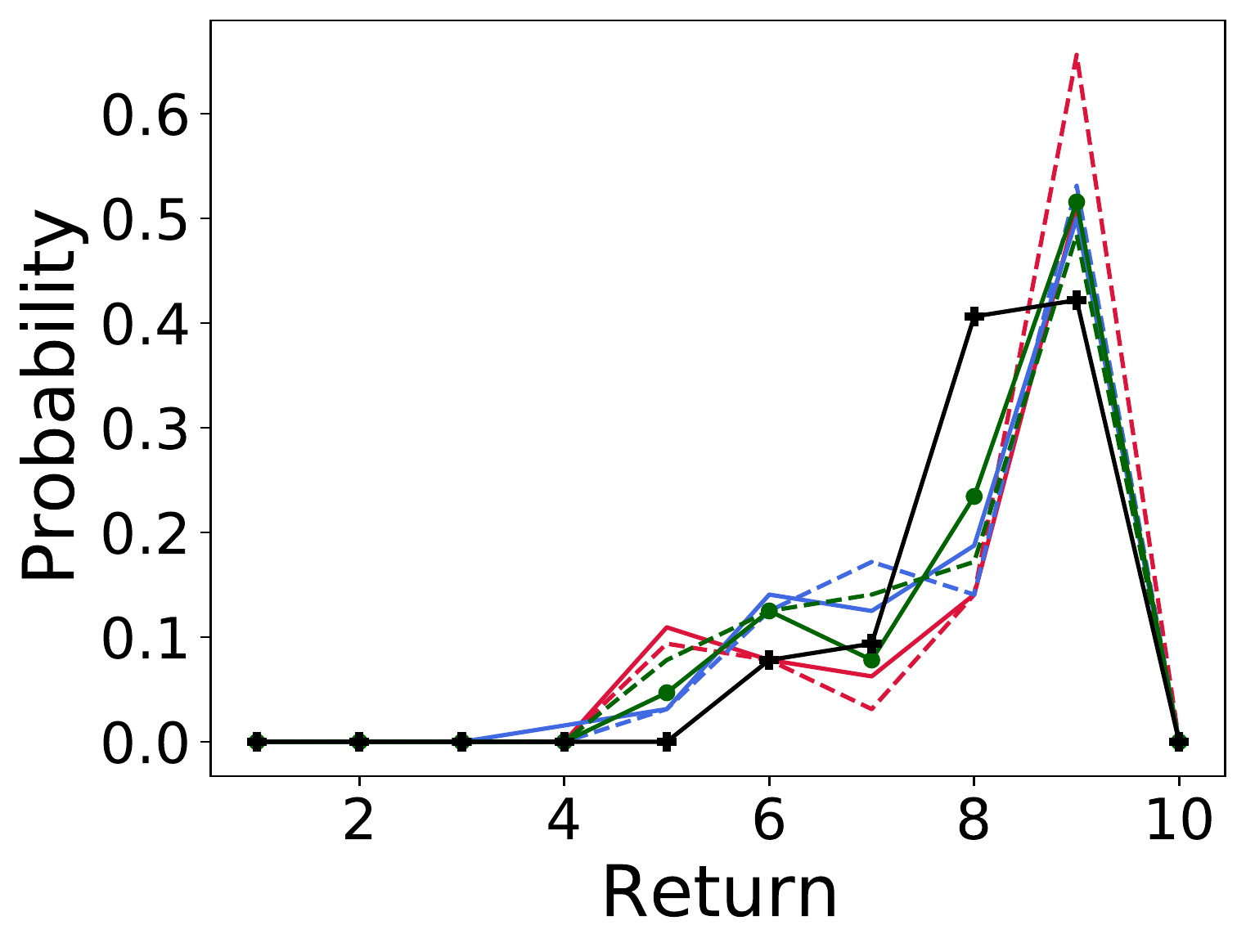}}
	\subfigure[\scriptsize $user_3$ ($s_0,a_0$)]{\includegraphics[width=0.272\textwidth]{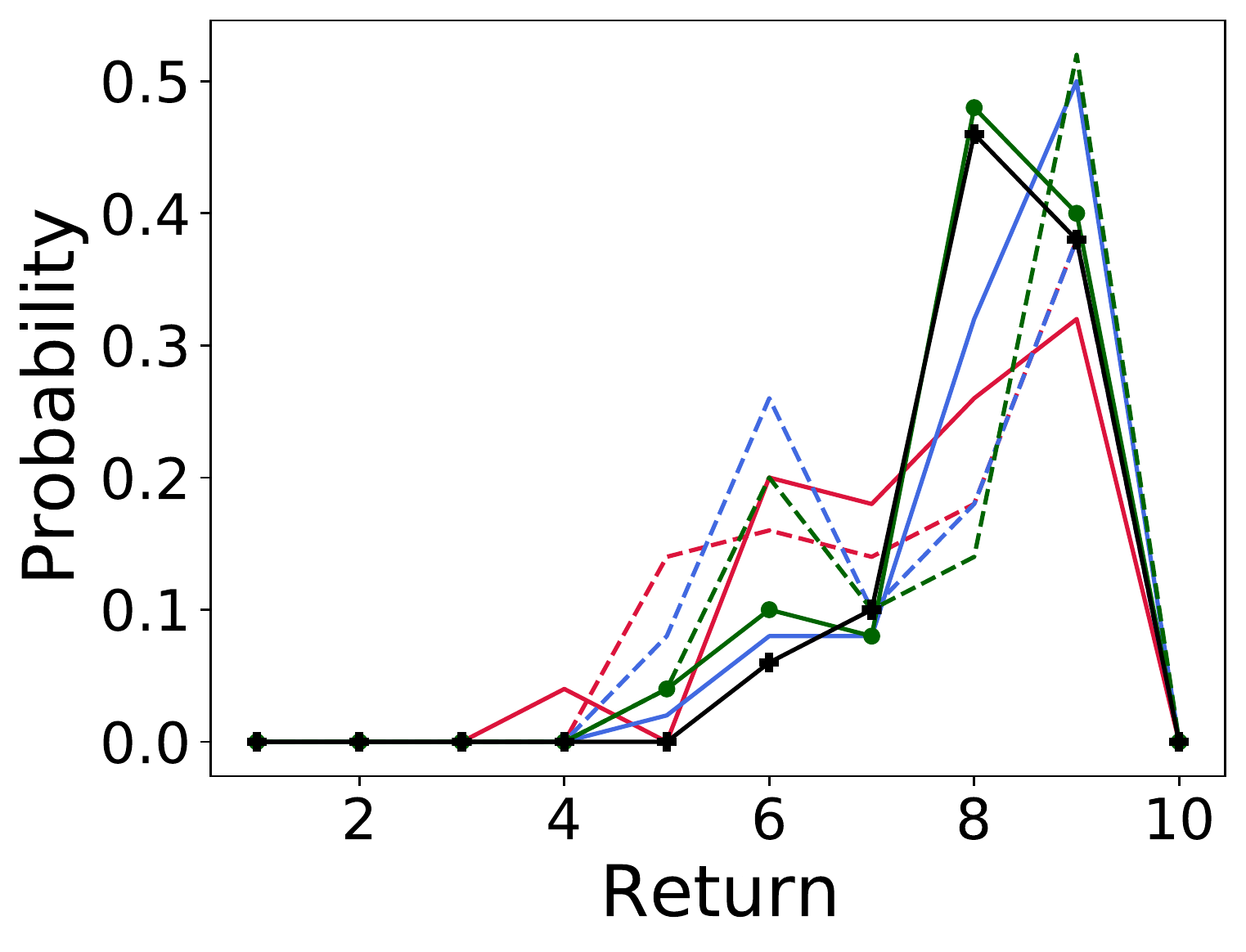}}
	\caption{The learned value distributions $Z^\pi(s_0,a_0)$ for various agents  given same initial state-action pair ($s_0,a_0$) for RecSim-2 agents. Initial ($s_0,a_0$) are randomly chosen, and returns are evaluated across randomly sampled 50 users from three different user types with discounted rewards over 20 steps with $\gamma=0.9$. \label{fig:frames}}
\end{figure*}

We pre-process the batch data to obtain incremental sessions, as used in \citep{wu2018session,liu2018stamp}. Each incremental session results in a tuple of $(s,a,r,s')$. 
The final data related statistics are available in Table \ref{tab:datastats}. We use 20$\%$ of the train set as hold-out validation set for tuning the hyperparameters in completely offline manner. 
Selecting the best hyperparameters using only offline logs and metrics while optimizing for the online evaluation metrics is a non-trivial task. This demands for the reliable offline evaluation metrics, preferably to be highly correlated with the online ones. 
We consider $R@3$ and $\Bar{Q}$ as offline evaluation metrics (described in Section \ref{sec:eval_metrics}). 
We find $\Bar{Q}$ to be a better metric in comparison to $R@3$. The performance of $\Bar{Q}$ is inline with the performance of online metrics as it is highly correlated with the online metrics, as shown in Table \ref{tab:pearson}. We also compare performance of BCQ and QRBCQ in terms of online and offline evaluation metrics while considering varying hyperparameters set, as shown in Fig \ref{fig:thresholding_appendix}. Results indicate the  reliability in considering $\Bar{Q}$ over $R@3$ for selecting best hyperparameters set. 
The finally selected best hyperparameters are summarized in Table \ref{tab:hyperparameters}.  
Further, the test environment for Diginetica (DN) is a bi-directional GRU of the same size as Q-Networks, and is trained to classify the response type for the item recommended by the agent given the items interacted so far.

\section{Comparison of learned value distributions of offline RL agents w.r.t. Online Policy \label{sec:app_e}}
We observe the learned value (return) distributions of various RAs. These RAs are trained using the logs generated by RecSim-2 behavior policy.
For this, we group the users having maximum interest in three randomly chosen categories as $User_1$, $User_2$ and $User_3$,  respectively. 
The returns are evaluated across randomly selected 50 users from each of the three user types with discounted rewards over $20$ steps with $\gamma$ = 0.9, given the same initial $(s_0,a_0)$ pair.
We compare these learned value distributions against the corresponding value distribution from the online policy using \textit{Wasserstein distance} \citep{dabney2018implicit}.
The Wasserstein distance related numbers are shown in Table \ref{tab:wasserstesin} whereas the respective learned value distributions are shown in Fig. \ref{fig:frames}. 
These results depict that the value distribution obtained using BCD4Rec agent is closer to the corresponding value distribution from the online agent in comparison to the value distributions obtained from other RAs.

\end{document}